\pdfoutput=1

\documentclass[11pt]{article}

\usepackage{EMNLP2022}

\usepackage{times}
\usepackage{latexsym}
\usepackage[T1]{fontenc}
\usepackage{tabularx}

\usepackage[utf8]{inputenc}

\usepackage{microtype}

\usepackage{inconsolata}

\usepackage{hyperref}
\usepackage{csquotes}
\usepackage{xspace}
\usepackage{graphicx}
\usepackage{multirow}
\usepackage{changepage,threeparttable} 
\usepackage{array,booktabs}
\usepackage[normalem]{ulem}
\useunder{\uline}{\ul}{}
\usepackage{siunitx}
\usepackage{multirow}
\usepackage{adjustbox}
\usepackage{amssymb}
\usepackage[normalem]{ulem}
\usepackage{pifont}
\newcommand{\xmark}{\ding{55}}%
\usepackage[inline]{enumitem}
\usepackage{amsmath}
\usepackage{xcolor}

\definecolor{atomictangerine}{rgb}{1.0, 0.6, 0.4}
\definecolor{cadmiumgreen}{rgb}{0.0, 0.42, 0.24}

\newcommand{\ignore}[1]{}
\newif\ifcomment
\commenttrue 
\ifcomment
\newcommand{\btocomment}[1]{\textcolor{red}{\bf \small [ #1 --BTO]}}
\newcommand{\micomment}[1]{\textcolor{violet}{\bf \small [ #1 --MI]}}
\newcommand{\agcomment}[1]{\textcolor{atomictangerine}{\bf \small [ #1 --AG]}}
\newcommand{\mkcomment}[1]{\textcolor{magenta}{\bf \small [ #1 --MK]}}

\newcommand{\kkcomment}[1]{\textcolor{blue}{\bf \small [ #1 --KK]}}
\newcommand{\wzcomment}[1]{\textcolor{blue}{\bf \small [ #1 --WZ]}}
\else
\newcommand{\btocomment}[1]{}
\newcommand{\micomment}[1]{}
\newcommand{\agcomment}[1]{}
\newcommand{\mkcomment}[1]{}
\newcommand{\kkcomment}[1]{}
\newcommand{\wzcomment}[1]{}
\fi

\newcommand{\ezcoref}{\texttt{ez{C}oref}\xspace}

\newcommand{\checkmarkcol}{\color{teal}\textbf{\checkmark}\color{black}}
\newcommand{\xmarkcol}{\color{red}\xmark\color{black}}

\title{\ezcoref: Towards Unifying Annotation Guidelines \\ for Coreference Resolution}

\author{Ankita Gupta$^{\spadesuit}$ \quad Marzena Karpinska$^{\spadesuit}$ \quad Wenlong Zhao$^\spadesuit$ \quad  Kalpesh Krishna$^{\spadesuit}$ \\ \bf \quad Jack Merullo$^\diamondsuit$ \quad Luke Yeh$^\heartsuit$  \quad Mohit Iyyer$^\spadesuit$ \quad Brendan O'Connor$^\spadesuit$\\\\
$^\spadesuit$University of Massachusetts Amherst, $^\diamondsuit$Brown University, $^\heartsuit$Google \\ \texttt{\{ankitagupta,mkarpinska,wenlongzhao,kalpesh,miyyer,brenocon\}@cs.umass.edu}\\ \texttt{john\_merullo@brown.edu} }

\begin{document}
\maketitle


\begin{abstract}

Large-scale, high-quality corpora are critical for advancing research in coreference resolution. 
However, existing datasets vary in their definition of coreferences and have been collected via complex and lengthy guidelines that are curated for linguistic experts.
These concerns have sparked a growing interest among researchers to curate a unified set of guidelines suitable for annotators with various backgrounds. In this work, we develop a crowdsourcing-friendly coreference annotation methodology, \ezcoref, consisting of an annotation tool and an interactive tutorial. We use \ezcoref\ to re-annotate 240 passages from seven existing English coreference datasets (spanning fiction, news, and multiple other domains) while teaching annotators only cases that are treated similarly across these datasets.\footnote{All resources accompanying this project will be added to our project page: \url{https://github.com/gnkitaa/ezCoref}} Surprisingly, we find that reasonable quality annotations were already achievable (>90$\%$ agreement between the crowd and expert annotations) even without extensive training. On carefully analyzing the remaining disagreements, we identify the presence of linguistic cases that our annotators unanimously agree upon but lack unified treatments (e.g., generic pronouns, appositives) in existing datasets. 
We propose the research community should revisit these phenomena when curating future unified annotation guidelines.

\end{abstract}

\section{Introduction}
\begin{figure}[t]
    \centering
  \includegraphics[width=0.45\textwidth, scale=0.7,clip,trim=0 0 0 0mm]{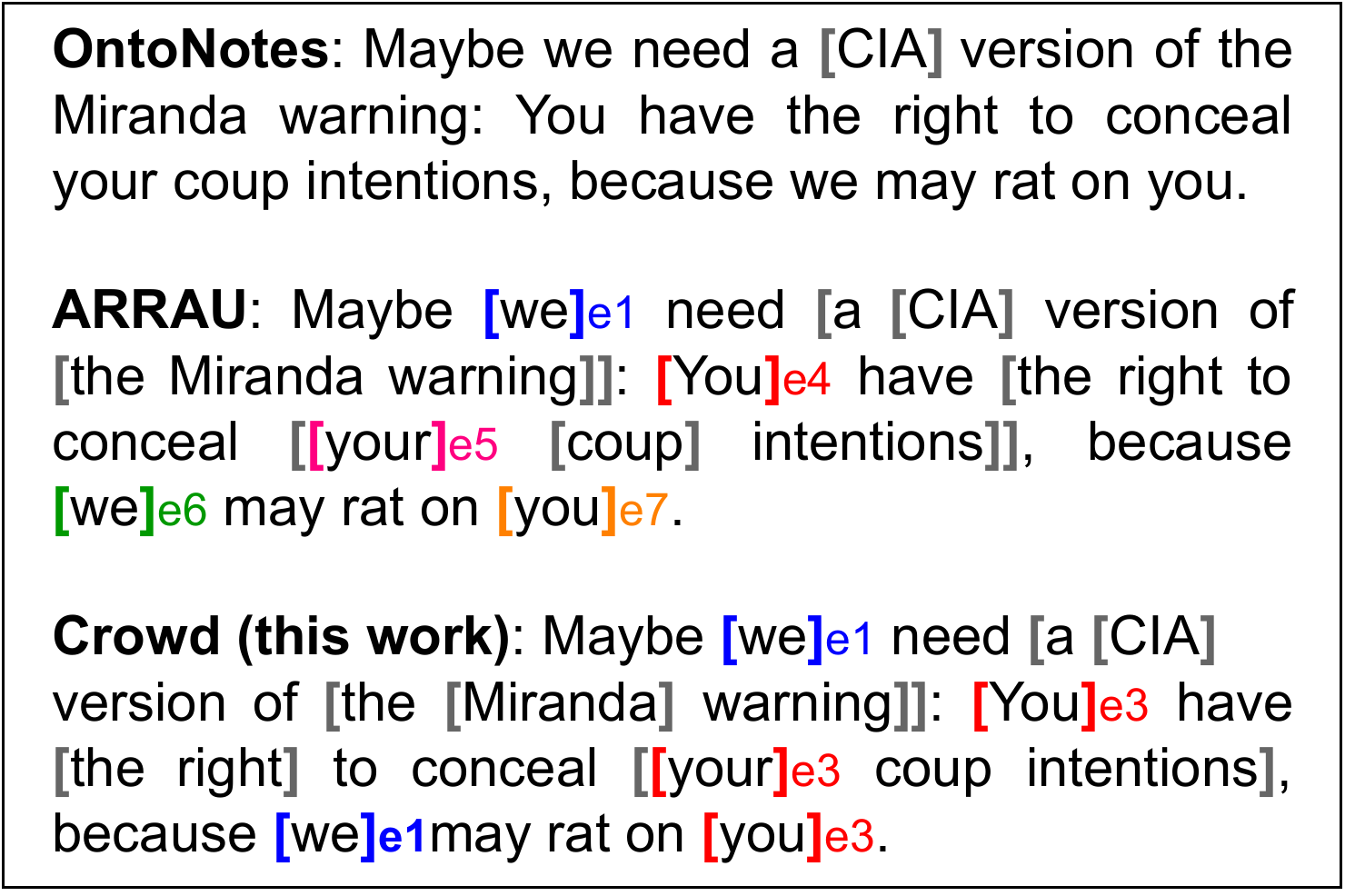}
  \caption{A common sentence from news domain annotated by two expert-curated datasets: OntoNotes~\cite{hovy-etal-2006-ontonotes} and ARRAU~\cite{Uryupina2019}. We visualize the two expert annotations along with the crowd annotations obtained in our study using \ezcoref\ platform. 
  While OntoNotes does not mark generic pronouns and ARRAU marks them as non-referring mentions, our crowdworkers mark all mentions of generic pronoun ``you” as coreferent and similarly for ``we”.
}
  \vspace{-0.1in}
\label{figure:data_guidelines}
\end{figure}

Coreference resolution is the task of identifying and clustering together all textual expressions (\emph{mentions}) that refer to the same discourse entity in a given document. Impressive progress has been made in developing coreference systems~\citep{lee-etal-2017-end, moosavi-strube-2018-using,
joshi-etal-2020-spanbert}, enabled by datasets annotated by experts
\citep{hovy-etal-2006-ontonotes, bamman-etal-2020-annotated, uryupina2020annotating} and crowdsourced  datasets~\citep{chamberlain-etal-2016-phrase}. However, these datasets vary widely in their definitions of coreference (expressed via annotation guidelines), 
resulting in inconsistent annotations both within and across domains and languages. For instance, as shown in Figure~\ref{figure:data_guidelines}, while  ARRAU~\cite{Uryupina2019} treats generic pronouns as non-referring, OntoNotes chooses not to mark them at all.

It is thus unclear which guidelines one should employ when collecting coreference annotations in a new domain or language.
Traditionally, existing guidelines have leaned towards lengthy explanations of complex linguistic concepts, such as those in the OntoNotes guidelines~\citep{ontonotes-guidelines}, which detail what should and should not be coreferent (e.g., how to deal with head-sharing noun phrases, premodifiers, and generic mentions). 
As a result, coreference datasets have traditionally been annotated by linguists (experts) already familiar with such concepts, which makes the process expensive and time-consuming. Crowdsourced coreference data collection has the potential to be significantly cheaper and faster; however, teaching an exhaustive set of linguistic guidelines to non-expert crowd workers remains a formidable challenge. As a result, there has been a growing interest among researchers in curating a unified set of guidelines~\cite{universal-anaphora} suitable 
for annotators with various backgrounds. 

More recently, games-with-a-purpose (GWAPs)~\cite{von-ahn-2006-gwap, poesio-etal-2013-PDoverview} were proposed to aid crowdsourcing of large coreference datasets~\cite{chamberlain-etal-2016-phrase-custom, yu-etal-PDv3-2022}.
While GWAPs make it enjoyable for crowdworkers 
to learn complex guidelines 
and perform annotations using them~\cite{madge2019progression}, 
they also require significant effort to attract and maintain workers. For instance, Phrase Detectives Corpus 1.0 was collected over a span of six years~\cite{chamberlain-etal-2016-phrase-custom, poesio2013phrase, yu-etal-PDv3-2022},  
which motivates us to instead study coreference collection on more efficient payment-based platforms. 

Specifically, our work investigates the quality of crowdsourced coreference annotations when annotators are taught only simple coreference cases that are treated uniformly across existing datasets (e.g., pronouns). By providing only these simple cases, we are able to teach the annotators the concept of coreference, while allowing them to freely interpret cases treated differently across the existing datasets. This setup allows us to identify cases where our annotators disagree among each other, but more importantly cases where they unanimously agree with each other but disagree with the expert, thus suggesting cases that should be revisited by the research community when curating future unified annotation guidelines. 


Our main contributions are:
\begin{enumerate}
    \item We develop a crowdsourcing-friendly coreference annotation methodology---\ezcoref--- 
    which includes an intuitive, open-sourced annotation tool
    supported by a short crowd-oriented interactive tutorial.\footnote{Our tutorial received overwhelmingly positive feedback. One annotator commented that it was ``\textit{absolutely beautiful, intuitive, and helpful. Legitimately the best one I've ever seen in my 2 years on AMT! Awesome job.}" (\autoref{table:tutorial_feedback} in Appendix)}
    \item We use \ezcoref to re-annotate 240 passages from seven existing English coreference datasets on Amazon Mechanical Turk (AMT), and conduct a comparative analysis of crowd and expert annotations. 
    We find that high-quality annotations are already achievable from non-experts without extensive training (>90$\%$ B3~\cite{bagga-1998-algorithms} agreement between crowd and expert annotations).
    \item We further qualitatively analyze remaining disagreements among crowd and expert annotations and identify linguistic cases that crowd unanimously marks as coreferent but
    lack unified treatment in existing datasets (e.g., generic pronouns as shown in~Figure~\ref{figure:data_guidelines}). 
    Additionally, analyzing inter-annotator agreement among the crowd reveals that crowd exhibits higher agreement when annotating familiar texts (e.g., childhood stories or fiction) compared to texts rich in cataphoras or those requiring world knowledge. 
    Finally, our qualitative analysis also provides an empirical evidence to support previous findings in literary studies (\citeauthor{szakolczai2016novels}’s \citeyearpar{szakolczai2016novels} analysis of Bleak House) and psychology (\citeauthor{Orvell2020}’s \citeyearpar{Orvell2020} claims about generic “you”).
\end{enumerate}

\begin{table*}[]
\centering 
\begin{adjustbox}{max width=\textwidth}
\tiny
\begin{tabular}{lccccclcccc}
\toprule
\multirow{2}{*}{\textbf{Dataset}} & \multirow{2}{*}{\textbf{\begin{tabular}[c]{@{}c@{}}Domains\\ $\#$(doc, ment, tok)\end{tabular}}} & \multirow{2}{*}{\textbf{Annotators}} & \multirow{2}{*}{\textbf{\begin{tabular}[c]{@{}c@{}}Mention \\ Detection\end{tabular}}} & \multicolumn{2}{c}{\textbf{Mention Types}} & \multicolumn{1}{c}{} & \multicolumn{3}{c}{\textbf{Coreference Links}} &  \\ \cline{5-6} \cline{8-11} \addlinespace[0.05cm]
 &  &  &  & \textbf{Singletons} & \textbf{\begin{tabular}[c]{@{}c@{}}Entity\\ Restrictions\end{tabular}} & \multicolumn{1}{c}{} & \textbf{Copulae} & \textbf{Appositives} & \textbf{Generics} & \textbf{Ambiguity} \\\midrule
 
\textbf{\begin{tabular}[c]{@{}l@{}}ARRAU\\ \cite{Uryupina2019}\end{tabular}} & \begin{tabular}[c]{@{}c@{}}Multiple\\ (552, 99K, 350K)\end{tabular}  & Single Expert & Manual & Yes & None &  & Special Link & No Link & Yes & Explicit \\ \addlinespace[0.2cm]

\textbf{\begin{tabular}[c]{@{}l@{}}Phrase Detectives (PD)\\ \cite{chamberlain-etal-2016-phrase}\end{tabular}}& \begin{tabular}[c]{@{}c@{}}Multiple\\ (542, 100K, 400K)\end{tabular} & \begin{tabular}[c]{@{}c@{}}Crowd (gamified) +\\ 2 Experts\end{tabular} & \begin{tabular}[c]{@{}c@{}}Semi\\ Automatic\end{tabular} & Yes & None &  & Special Link & Special Link & Yes & Implicit \\ \addlinespace[0.2cm]

\textbf{\begin{tabular}[c]{@{}l@{}}GUM\\\cite{zeldes-2017-gum}\end{tabular}} & \begin{tabular}[c]{@{}c@{}}Multiple\\ (25, 6K, 20K)\end{tabular}  & \begin{tabular}[c]{@{}c@{}} Experts \\ (Linguistics Students)\end{tabular} & Manual & Yes & None &  & \begin{tabular}[c]{@{}c@{}} Coref \\ (Sub-Types)  \end{tabular} & \begin{tabular}[c]{@{}c@{}} Coref \\ (Sub-Type)  \end{tabular} & Yes & None \\ \addlinespace[0.2cm]

\textbf{\begin{tabular}[c]{@{}l@{}}PreCo \\ \cite{chen-etal-2018-preco}\end{tabular}} & \begin{tabular}[c]{@{}c@{}}Multiple***\\ (38K, 3.58M, 12.5M)\end{tabular}  & \begin{tabular}[c]{@{}c@{}}Non-Expert, \\ Non-Native \end{tabular}& Manual** & Yes & None &  & Coref & Coref & Yes & None \\ \addlinespace[0.1cm]

\textbf{\begin{tabular}[c]{@{}l@{}}OntoNotes\\ \cite{hovy-etal-2006-ontonotes}\end{tabular}} & \begin{tabular}[c]{@{}c@{}}Multiple\\ (1.6K, 94K, 950K)\end{tabular}  & Experts & Mixed & No & None &  & Special Link & Special Link & \begin{tabular}[c]{@{}c@{}}Only with\\ Pronominals\end{tabular} & None \\ \addlinespace[0.2cm]

\textbf{\begin{tabular}[c]{@{}l@{}}LitBank\\ \cite{bamman-etal-2020-annotated}\end{tabular}} & \begin{tabular}[c]{@{}c@{}}Single\\ (100, 29K, 210K)\end{tabular}  & Experts & Manual & Yes & ACE (selected) &  & Special Link & Special Link & \begin{tabular}[c]{@{}c@{}}Only with\\ Pronominals\end{tabular} & None \\ \addlinespace[0.2cm]

\textbf{\begin{tabular}[c]{@{}l@{}}QuizBowl\\ \cite{guha-etal-2015-removing}\end{tabular}} & \begin{tabular}[c]{@{}c@{}}Single\\ (400, 9.4K, 50K)\end{tabular}  & \begin{tabular}[c]{@{}c@{}}Domain\\ Experts\end{tabular} & \begin{tabular}[c]{@{}c@{}}Manual\\ \& CRF*\end{tabular} & Yes & \begin{tabular}[c]{@{}c@{}}Characters,\\ Books,\\  Authors*\end{tabular} &  & Coref & Coref & If Applicable & None \\ \midrule
\textbf{\begin{tabular}[c]{@{}l@{}}ezCoref Pilot Dataset 
\\ (this work)\end{tabular}} & Multiple & Crowd (paid) & \begin{tabular}[c]{@{}c@{}}Fully\\ Automatic\end{tabular} & Yes & None &  & \begin{tabular}[c]{@{}c@{}}Annotator`s\\ Intuition\end{tabular} & \begin{tabular}[c]{@{}c@{}}Annotator`s\\ Intuition\end{tabular} & \begin{tabular}[c]{@{}c@{}}Annotator`s\\ Intuition\end{tabular} & Implicit \\ \bottomrule
\end{tabular}
\end{adjustbox}

\caption{Summary of seven datasets analyzed in this work,
which 
differ in domain, size, annotator qualifications, mention detection procedures, types of mentions, and types of links considered as coreferences between these mentions.\scriptsize{*Allows other types of mention only when this mention is an answer to a question.}\scriptsize{**We interpret manual identification based on illustrations presented in the original publication~\cite{chen-2018-preco}. }\scriptsize{***See Footnote \ref{fn:preco}.}} 
\vspace{-0.1in}
\label{table:motivation}
\end{table*}

\section{Related Work}
\label{sec:related_work}

\paragraph{Existing coreference datasets:} 
\autoref{table:motivation} provides an overview of seven prominent coreference datasets, which differ widely in their annotator population, mention detection, and coreference guidelines.\footnote{Many others exist too; for example, see Jonathan Kummerfeld's \href{https://docs.google.com/spreadsheets/d/1DTuzCAOcqEzKqQNFcLVAx0KQb1DTUYVQbZV55LONXHI/edit\#gid=0}{spreadsheet list} (accessed Jan.\ 2022).} 
Many datasets are annotated by experts heavily trained in linguistic standards, including ARRAU~\cite{Uryupina2019}, LitBank~\cite{bamman-etal-2020-annotated},
GUM~\cite{zeldes-2017-gum}, and OntoNotes~\cite{hovy-etal-2006-ontonotes}). 
Due to its scale and quality, OntoNotes is likely the most widely used for NLP coreference research, including in two CoNLL shared tasks \cite{pradhan-etal-2011-conll,pradhan-etal-2012-conll}. Coreference datasets annotated by non-experts include
those created by part-time non-native English speakers (PreCo; \citet{chen-etal-2018-preco}),
domain but not linguistic experts (QuizBowl; \citet{guha-etal-2015-removing}), and gamified crowdsourcing without financial compensation (Phrase Detectives; \citet{chamberlain-etal-2016-phrase-custom}). 

\paragraph{Coreference annotation tools:}


Several coreference annotation tools with similar features to \ezcoref\ have already been developed (See \autoref{tab:tools-prior-work-compare} in Appendix for more details). However, these are difficult to port to a crowdsourced workflow, as they require users to install software on their local machine~\cite{widlcher-2012-glozz, landragin-etal-2012-analec, kopec-2014-mmax2, reiter-2018-corefannotator}, or have complicated UI design with multiple drag and drop actions and/or multiple windows~\cite{stenetorp-etal-2012-brat-custom, widlcher-2012-glozz, landragin-etal-2012-analec, yimam-etal-2013-webanno, girardi-etal-2014-cromer, kopec-2014-mmax2, oberle-2018-sacr}. Closest to our work is \texttt{CoRefi}~\citep{bornstein-2020-corefi}, a web-based coreference annotation tool that can be embedded into crowdsourcing websites.  Subjectively, we found its user interface difficult to use (e.g., users have to memorize multiple key combinations). It also does not allow for nested spans, reducing its usability. 

\paragraph{Crowdsourcing coreference annotations:} 
Several efforts have been made to crowdsource linguistic annotations~\cite{snow-2008-cheap, callison-burch-2009-fast, howe2008crowdsourcing, lawson-etal-2010-annotating}, including on payment-based microtasks via platforms like AMT and GWAPs~\cite{von-ahn-2006-gwap}. Many GWAPs~\cite{poesio-etal-2013-phrase, kicikoglu2019wormingo, madge-et-al-2019-wordclicker, fort-et-al-2014-zombilingo} have been used in NLP to collect linguistic annotations including coreferences; with some broader platforms~\cite{venhuizen-etal-2013-gamification, madge2019progression} aiming to gamify the entire text annotation pipeline. One solution to teaching crowd workers complex guidelines is to incorporate \textit{learning by progression}~\cite{kicikoglu2020aggregation, madge2019progression, miller2019expertise}, where annotators start with simpler tasks and gradually move towards more complex problems, but this requires subjective judgments of task difficulty. In contrast to the payment-based microtask setting studied in this work, GWAPs are not open-sourced, need significant development, take longer to collect data, and require continuous efforts to maintain visibility~\cite{poesio-etal-2013-phrase}.
\section{\ezcoref: A Crowdsourced Coreference Annotation Platform}
\label{sec:software}

The \ezcoref user experience consists of (1) a step-by-step interactive tutorial 
and (2) an annotation interface, which are part of a pipeline including automatic mention detection and AMT Integration.



\paragraph{Annotation structure:}
Two annotation approaches are prominent in the literature: (1) a local pairwise approach, 
annotators are shown a pair of mentions and asked whether they refer to the same entity \cite{hladka-etal-2009-play, chamberlain-etal-2016-phrase, li-etal-2020-active-learning, ravenscroft-etal-2021-cd}, which is time-consuming;
or (2) a cluster-based approach~\cite{reiter-2018-corefannotator, oberle-2018-sacr, bornstein-2020-corefi}, in which annotators group all mentions of the same entity into a single cluster. In \ezcoref\ we use the latter approach, which can be faster but requires the UI to support more complex actions for creating and editing cluster structures.
\begin{figure}[t]
    \centering
  \includegraphics[width=0.48\textwidth, scale=1,clip,trim=0 0 0 0mm]{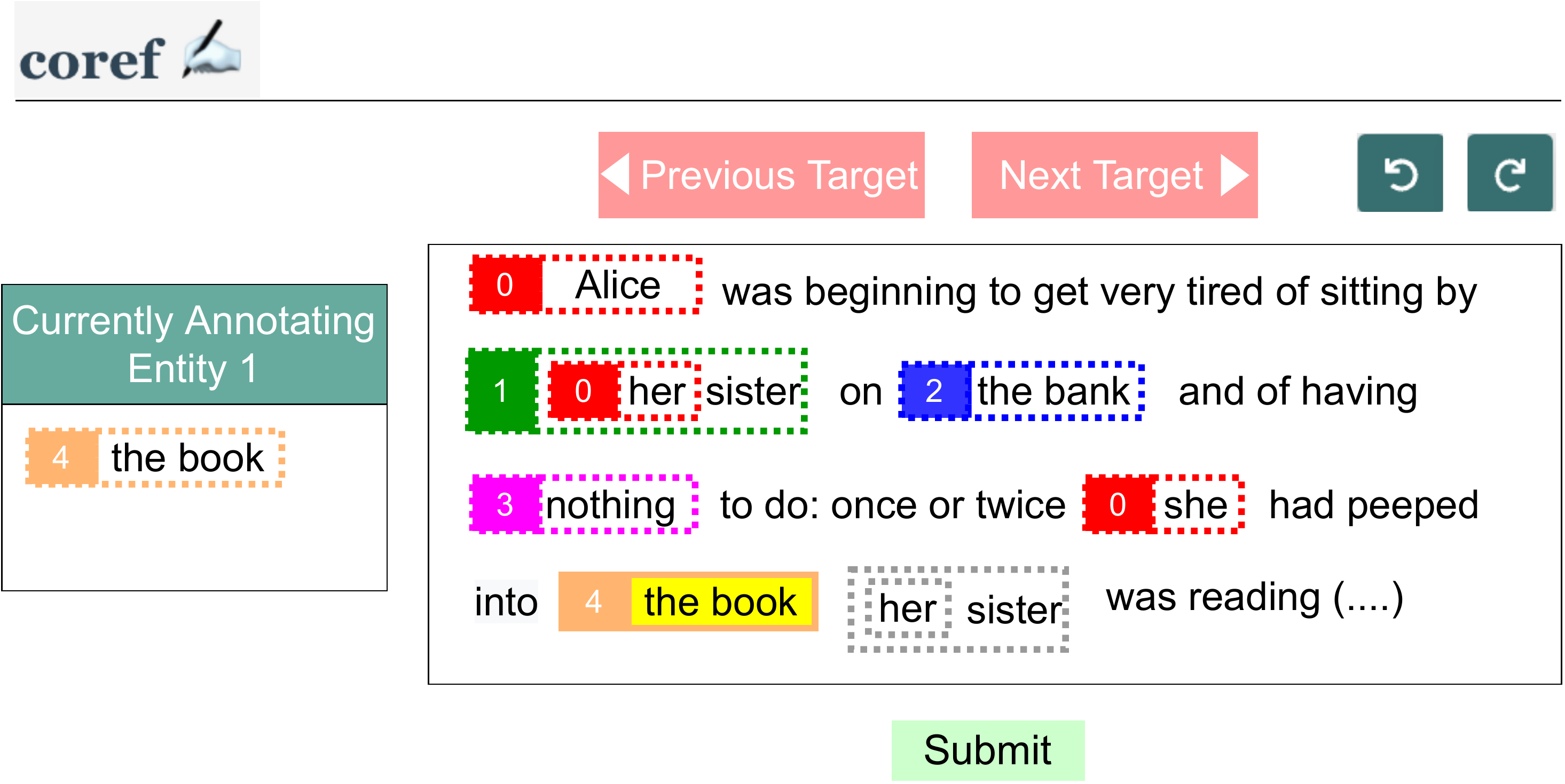}
  \caption{Part of the \ezcoref interface (\S\ref{sec:software})}
  \label{figure:interface}
\end{figure}
\paragraph{User interface:}
We spent two years iteratively designing, implementing, and user testing the interface to make it as simple and crowdsourcing-friendly as possible (\autoref{figure:interface}).\footnote{The interface is implemented in \href{https://reactjs.org/}{ReactJS}.}
Marked mentions are surrounded by color-coded frames with entity IDs. 
The currently selected mention (``the book"),
is highlighted with a flashing yellow cursor-like box.
The core annotation action is to select other mentions that corefer with the current mention, and then advance to a later unassigned mention; annotators can also re-assign a previously annotated mention to another cluster.  
Advanced users can exclusively use keyboard shortcuts,
undo and redo actions were added to allow error correction.
Finally, \ezcoref provides a side panel showing mentions of the entity currently being annotated to spot mentions assigned to the wrong cluster.


\paragraph{Coreference tutorial:}
To teach crowdworkers the basic definition of coreference and familiarize them with the interface, we develop a tutorial (aimed to take $\sim 20$ minutes) that familiarizes them with the mechanics of the annotation tool, and then trains them in a minimal set of annotation guidelines (\autoref{table:tutorial_example}). Our minimal guidelines cover cases that are annotated similarly across most guidelines and are unlikely to be disputed.
The tutorial concludes with a quality control example to exclude poor quality annotators.\footnote{Examples of the tutorial interface and the quality control example are provided in Appendix.} Training examples, feedback, and annotation guidelines can be easily customized using a simple JSON schema. 

\paragraph{Annotation workflow:}
The annotators are presented with one passage (or ``document'') at a time
(\autoref{figure:interface}), and all mentions have to be annotated before proceeding to the next passage. There is no limitation to the length or language of the passage.
In this work, we divide an initial document into a sequence of shorter passages of complete sentences, on average 175 tokens, as shorter passages minimize the need to scroll, reducing annotator effort.  
While this obviously cannot capture longer distance coreference,\footnote{We leave this for future work---for example, more sophisticated user interfaces to support longer documents, or merging coreference chains between short passages. As documents get progressively longer, such as book chapters or books, the task takes on aspects of cross-document coreference and entity linking\ (e.g.\ \citealp{bagga-baldwin-1998-entity-based,fitzgerald-etal-2021-moleman, logan-iv-etal-2021-benchmarking}).}
a large portion of important coreference phenomena is local:
within the OntoNotes written genres, for pronominal mentions,
the closest antecedent is contained within the current or previous two sentences more than 95\% of the time.


\begin{table}[]
\centering \tiny
\begin{tabular}{ll}
\toprule
\textbf{Example} &
  \textbf{Phenomena Taught} \\ \hline
\begin{tabular}[c]{@{}l@{}}\textcolor{red}{{[}John{]}} doesn’t like \textcolor{teal}{{[}Fred{]}}, but \textcolor{red}{{[}he{]}} still \\ invited \textcolor{teal}{{[}him{]}}to \textcolor{violet}{{[}the party{]}}.\end{tabular} &
  \begin{tabular}[c]{@{}l@{}} (1) personal pronouns\\ (2) singletons\end{tabular} \\
  \addlinespace[0.2cm]
  
\begin{tabular}[c]{@{}l@{}}\textcolor{teal}{{[}This dog{]}} likes to play \textcolor{blue}{{[}catch{]}}.\textcolor{teal}{{[}It{]}}’s \\ better than other \textcolor{magenta}{{[}dogs{]}} at \textcolor{blue}{{[}this game{]}}. \\ \textcolor{red}{{[}\textcolor{teal}{{[}Its{]}} owner{]}} is really proud.\end{tabular} &
  \begin{tabular}[c]{@{}l@{}}(1) possessive pronouns\\ (2) semantically similar expression \\ which are not coreferring \\ (3) non-person entities (animals) \end{tabular} \\ \addlinespace[0.2cm]
  
\begin{tabular}[c]{@{}l@{}}\textcolor{teal}{{[}Director \textcolor{magenta}{{[}Mackenzie{]}}{]}} spent \textcolor{red}{{[}last two years{]}}\\  working on a \textcolor{blue}{{[}“Young Adam”{]}}. During \\ \textcolor{red}{{[}this time{]}} \textcolor{teal}{{[}he{]}} often had to make \textcolor{violet}{{[}compromises{]}}\\ but \textcolor{blue}{{[}the movie{]}} turned out to exceed expectations.\end{tabular} &
  \begin{tabular}[c]{@{}l@{}} (1) nested spans \\ (2) non-person entities (time, item) \end{tabular} \\ \addlinespace[0.2cm]
  
\begin{tabular}[c]{@{}l@{}}\textcolor{teal}{{[}The office{]}} wasn’t exactly small either.\\ \textcolor{red}{{[}I{]}}’m sure that 50, or maybe even 60,  \textcolor{violet}{{[}people{]}}\\ could easily fit \textcolor{teal}{{[}there{]}}.\end{tabular} &
  \begin{tabular}[c]{@{}l@{}} (1) non-person entities (place) \end{tabular} \\ \bottomrule
\end{tabular}
\caption{Minimal guidelines explained by our tutorial.}
\label{table:tutorial_example}
\end{table}

\paragraph{Automatic mention detection:}
\label{sec:mention_detection}
As a first step to collect coreference annotations, we must identify mentions in the documents from each of the seven existing datasets; this process is done in a diverse array of ways (from manually to automatic) in prior work as shown in \autoref{table:motivation}. We decided to automatically identify mentions to give all crowdworkers an identical set of mentions, which simplifies the annotation task and also allows us to easily compare and study their coreference annotations via inter-annotator agreement.
Specifically, we implement a simple algorithm that yields a high average recall over all seven datasets.\footnote{Note that any mention detection algorithm can be used as long as its recall across all datasets is high, and ours is only one such algorithm. However, we don’t collect and compare crowd annotations for mentions obtained from these potential algorithms as it would be prohibitively expensive. Furthermore, while advanced mention detection methods can definitely improve annotation quality, our goal is not to collect highest quality coreference dataset, but to study annotator behaviour when provided a common set of mentions.}

It considers all noun phrases (including proper nouns, common nouns, and pronouns) as markables, extracting them using the Stanza dependency parser (version 1.3.0; ~\citet{qi2020stanza}). We allow for nested mentions and proper noun premodifiers (e.g.,  \textcolor{red}{\textbf{[}}U.S.\textcolor{red}{\textbf{]}}  in ``U.S. policy''). We also include all conjuncts with the entire coordinated noun phrase (\textcolor{red}{\textbf{[}}Mark\textcolor{red}{\textbf{]}}, \textcolor{red}{\textbf{[}}Mary\textcolor{red}{\textbf{]}}, as well as \textcolor{red}{\textbf{[}}Mark and Mary\textcolor{red}{\textbf{]}}, are all considered mentions);  see Appendix~\ref{sec:md_algo} for more details.

\section{Using \ezcoref\ to Re-annotate Existing Coreference Datasets}
\label{sec:data_collection}

We deploy \ezcoref\ on the AMT crowdsourcing platform to re-annotate 240 passages from seven existing datasets, covering seven unique domains. In total, we collect annotations for 12,200 mentions and 42,108 tokens. We compare our workers' annotations both quantitatively and qualitatively to each other and to existing expert annotations.





\paragraph{Datasets:}  
We collect coreference annotations for the seven existing datasets described in Table~\ref{table:motivation}: OntoNotes~\cite{hovy-etal-2006-ontonotes}, LitBank~\cite{bamman-etal-2020-annotated}, PreCo\footnote{The PreCo dataset is interestingly large but seems difficult to access. In November 2018 and October 2021 we filled out the data request form at the URL provided by the paper,
and attempted to contact the PreCo official email directly,
but did not receive a response. To enable a precise research comparison, we scraped all documents from PreCo's public demo in November 2018 (no longer available as of 2021); its statistics match their paper and our experiments use this version of the data.
PreCo further suffers from data curation issues \cite{gebru_documentation, Jo2020}; it uses text from English reading comprehension tests collected from several websites,
but the original document sources and copyright statuses
are undocumented. When reading through PreCo documents, we found many domains including opinion, fiction, biographies, and news (\autoref{table:datasets-description-detailed} in Appendix); we use our manual categories for domain analysis. \label{fn:preco}}~\cite{chen-etal-2018-preco}, ARRAU~\cite{Uryupina2019}, GUM~\cite{zeldes-2017-gum}, Phrase Detectives~\cite{chamberlain-etal-2016-phrase}, and QuizBowl~\cite{guha-etal-2015-removing}. The sample covers seven domains: news, opinionated magazines, weblogs, fiction, biographies, Wikipedia articles, and trivia questions from Quiz Bowl. 
For each dataset with multiple domains, we manually select domain(s) to re-annotate so that we cover a broad range of domains. From each domain in each dataset, we then select documents and divide them into shorter passages (on average 175 tokens each), creating 20 such passages per dataset. For datasets with multiple domains, we choose 20 such passages per domain (see Appendix~\ref{sec:crowd_data} for detail). Overall, we collect annotations for 240 passages with 5 annotations per passage to measure inter-annotator agreement. 

\paragraph{Procedure:}
We first launch an annotation tutorial (paid \$4.50) and recruit the annotators on the AMT platform.\footnote{We allow only workers with a >= 99\% approval rate and at least 10,000 approved tasks who are from the US, Canada, Australia, New Zealand, or the UK.} At the end of the tutorial, each annotator is asked to annotate a short passage (around 150 words). 
Only annotators with a B3 score~\cite{bagga-1998-algorithms} of \num{0.90} or higher are then invited to participate in the annotation task. 

\paragraph{Training Annotators with Minimal Guidelines using \ezcoref:}
\begin{figure}[th]
\centering
\includegraphics[width=0.4\textwidth, scale=0.5,clip,trim=0 0 0 0mm]{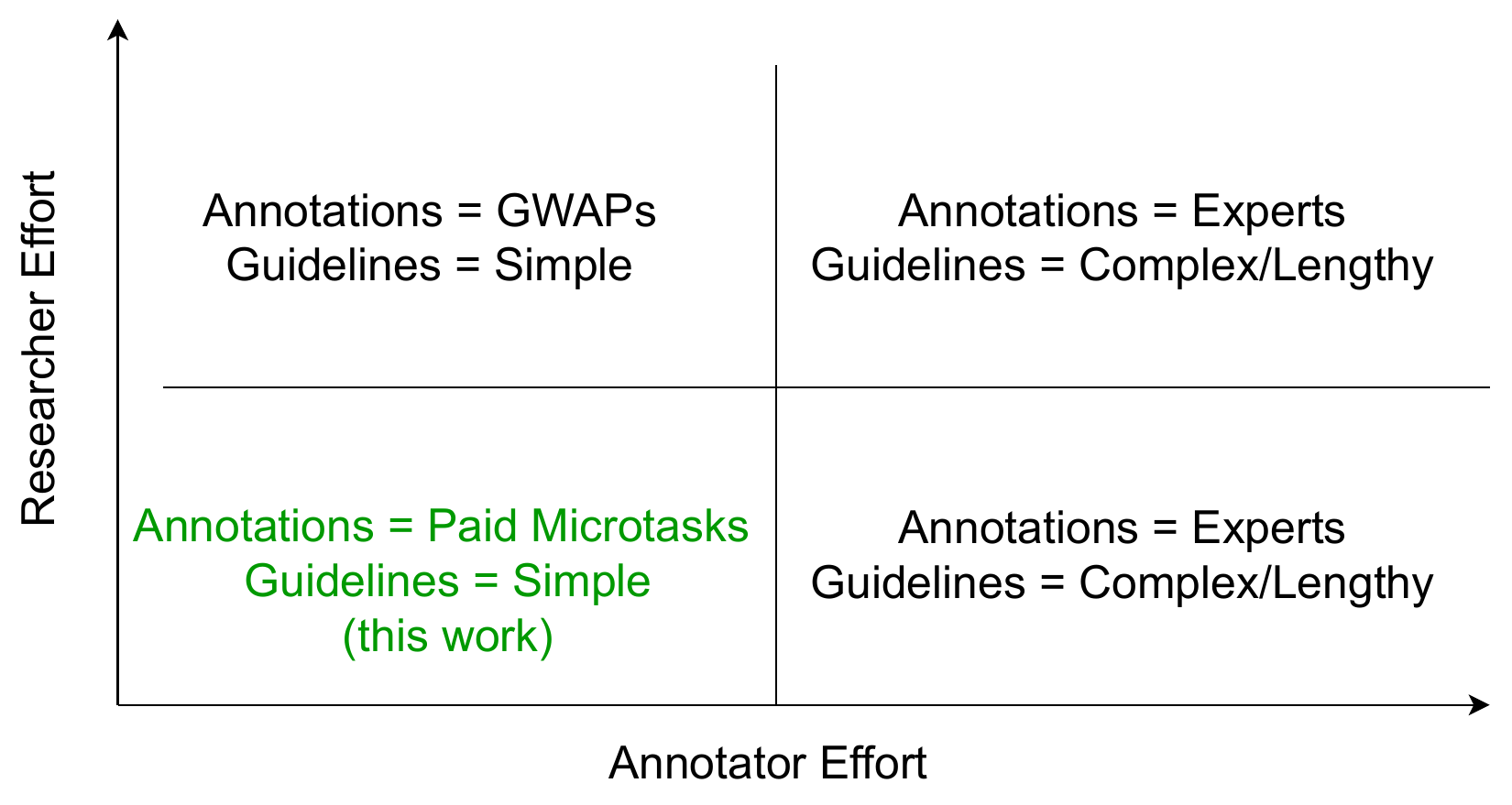}
\caption{
Existing expert annotated datasets entail high annotator effort (e.g., OntoNotes, ARRAU). Existing crowdsourced coreference datasets (e.g., Phrase Detectives) entail significant researcher effort. In this work, we explore the minimum effort scenario for both annotators (by providing them minimal guidelines) and researchers (by open-sourcing \ezcoref).} 
\label{figure:motivation_efforts}
\end{figure}
As the goal of our study is to understand what crowdworkers perceive as coreference and to identify instances of genuine ambiguity, we train our annotators with minimal guidelines. We carefully draft our training examples to include only cases which are considered as coreference by all the existing datasets. The objective is to teach crowdworkers the broad definition of coreference while leaving space for different interpretations of ambiguous cases or those resolved differently across the existing datasets. Note that a comparable experiment with more complex guidelines is infeasible since it is unclear which guidelines to choose, and also providing complex linguistic guidelines to crowdworkers remains an open challenge.
Overall, \ezcoref\ is aimed to minimize both researcher and annotator effort for new coreference data collection, in comparison to prior work (\autoref{figure:motivation_efforts}).



\paragraph{Worker details:}
Overall, 73 annotators (including 44 males, 20 females, and one non-binary person)\footnote{We did not collect demographic data for the remaining eight individuals, from an earlier pilot experiment.} completed the tutorial task, which took 19.4 minutes on average (sd=11.2 minutes). They were aged between 21 and 69 years (mean=38.9, sd=11.3) and identified themselves as native English speakers. 
Most of the annotators had at least a college degree (47 vs 18). 89.0$\%$ of annotators, who did the tutorial, received a B3 score of 0.90 or higher for the final screening example, and were invited to the annotation task. 50.7$\%$ of the invited annotators returned to participate in the main annotation task, and 29.2\% of them annotated five or more passages. Annotation of one passage took, on average, 4.15 minutes,
a rate of 2530 tokens per hour. 
The total cost of the tutorial was \$460.70. We paid \$1 per passage for the main annotation task, resulting in a total cost of \$1440.\footnote{All reported costs include 20$\%$ AMT fee.}

\section{Analysis}
\label{sec:data_analysis}
In this section, we perform quantitative and qualitative analyses of our crowdsourced coreference annotations. First, we evaluate the performance of our mention detection algorithm, comparing it to gold mentions across seven datasets. Next, we measure the quality of our annotations (via inter-annotator agreement between our crowdworkers) and their agreement with other datasets. Finally, we discuss interesting qualitative results.  
\subsection{Mention Detector Evaluation}

Datasets differ in the way they define their mention boundaries. Hence, the boundaries for the same mention may differ. To fairly compare our mentions with the gold standards, we employ a headword-based comparison. We find the head of the given phrase by identifying, in the dependency tree, the most-shared ancestor of all tokens within the given mention. Two mentions are considered same if their respective headwords match.

\autoref{table:mention_detection_algo} compares our mention detector to the gold mentions in existing datasets. Our method obtains high recall across most datasets ($>$\num{0.90}), which shows that most of the mentions annotated in existing datasets are correctly identified and allows a direct comparison of crowd annotations with expert annotations.
It has the lowest recall with ARRAU  (\num{0.84}) and PreCo (\num{0.88}), which is to be expected as ARRAU marks all referring premodifiers (identified manually) and PreCo allows common noun modifiers, while we identify only the premodifiers which are proper nouns.\footnote{We made this decision as identifying automatically all premodifiers would result in many singletons and lead to more arduous annotation effort.}
For most datasets, the precision is $>$\num{0.80}, suggesting that the algorithm identifies most of the relevant mentions. 
We observe a substantially lower score for OntoNotes, LitBank, and QuizBowl as these datasets restrict their mention types to limited entities (refer to \autoref{table:motivation}). However, low precision on these datasets is expected and does not affect our analysis since an algorithm with high precision on LitBank or OntoNotes would miss a huge percentage of relevant mentions and entities on other datasets (constraining our analysis) and when annotating new texts and domains. 
Moreover, our algorithm identifies more mentions than in the original datasets, which also allows us to discover new entities. 
Finally, the mention density (number of mentions per token) from our detector 
remains roughly consistent across all datasets when using our method, allowing us to fairly compare statistics (e.g., agreement rates) across datasets.

\begin{table}[]
\centering \scriptsize
\begin{tabular}{lcccc}
\toprule \addlinespace[0.1cm]
\multicolumn{1}{c}{\multirow{2}{*}{\textbf{Dataset}}} &
  \multirow{2}{*}{\textbf{Recall}} &
  \multirow{2}{*}{\textbf{Precision}} &
  \multicolumn{2}{c}{\textbf{Mentions / Tokens}} \\ \addlinespace[0.1cm] \cline{4-5} \addlinespace[0.1cm]
\multicolumn{1}{c}{}       &                      &                      & \textbf{Gold}        & \textbf{This Work}      \\ \hline \addlinespace[0.1cm]
OntoNotes                  & 0.957 & 0.376 & 0.112 & 0.286 \\
LitBank                    & 0.962                & 0.415                & 0.121                & 0.280                \\
QuizBowl                   & 0.956                & 0.543                & 0.188                & 0.318                \\
PD (Gold)   & 0.953                & 0.803                & 0.259                & 0.273                \\
PD (Silver) & 0.938                & 0.791                & 0.265                & 0.274                \\
GUM                        & 0.906                & 0.848                & 0.269                & 0.287                \\
PreCo                      & 0.881                & 0.883                & 0.287                & 0.287                \\
ARRAU                      & 0.840 & 0.870 & 0.289 & 0.279 \\

\bottomrule
\end{tabular}
\vspace{-0.05in}
\caption{Comparison of mentions identified by our mention detection algorithm with the gold mentions annotated in the respective datasets. We use head-word based comparison to compare mentions of different lengths.}
\vspace{-0.1in}
\label{table:mention_detection_algo}
\end{table}

\subsection{Agreement with Existing Datasets} 
\begin{figure}[]
    \centering
   \includegraphics[width=0.5\textwidth, scale=1,clip,trim=0 0 0 0mm]{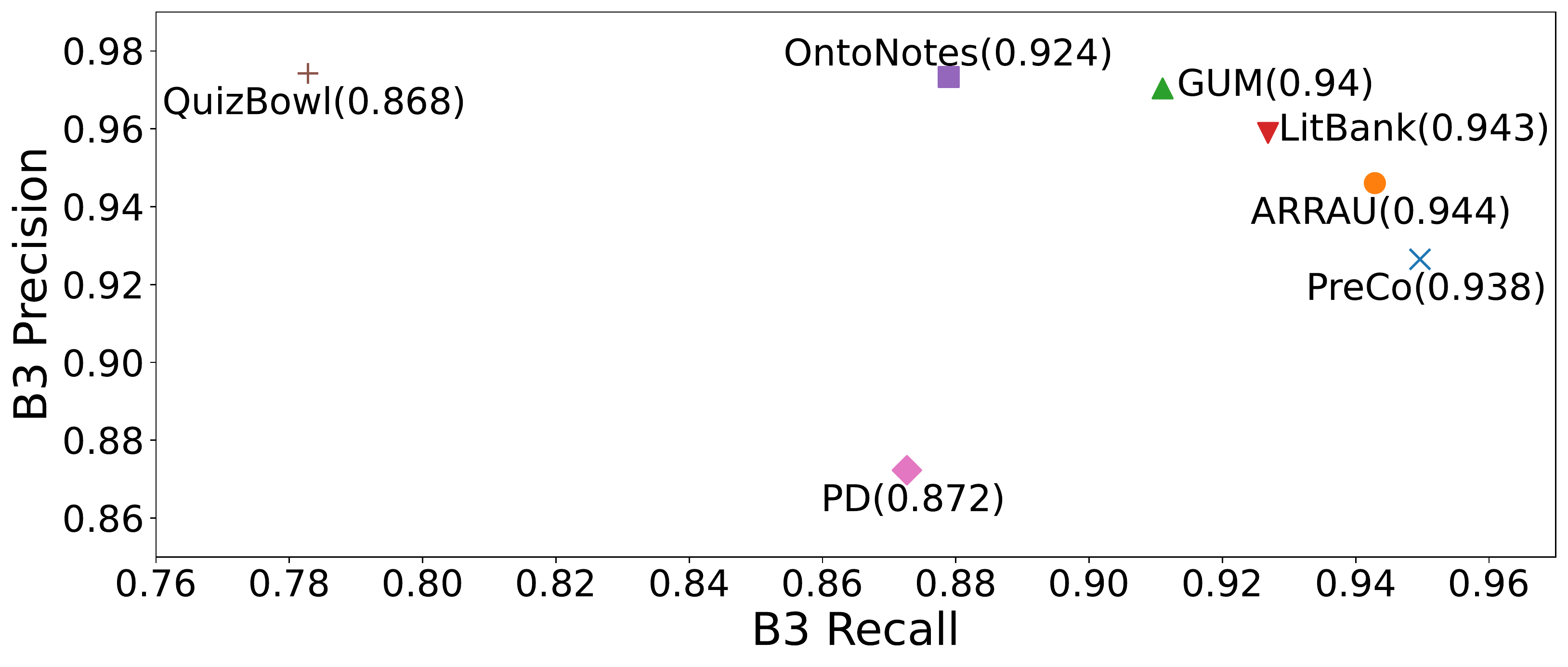}
  \caption{Agreement with gold annotations across datasets. B3 (F1) scores shown in parentheses are computed with singletons included.
  }
  \vspace{-0.1in}
  \label{fig:agreement_with_gold_data}
\end{figure}

How well do annotations from \ezcoref agree with annotations from existing datasets? 
\paragraph{Aggregating annotations:}
To compare crowdsourced annotations with gold annotations, we first require an aggregation method that can combine annotations from multiple crowdworkers to infer coreference clusters. We use a simple aggregation method that determines whether a pair of mentions is coreferent by counting the number of annotators who marked the two mentions in the same cluster.\footnote{Future data collection efforts interested in creating large resources can utilize more advanced aggregation methods~\cite{poesio2019crowdsourced}.} Two mentions are considered as coreferent when the number of annotators linking them together is greater than a threshold ($\tau$). After inferring these pairs of mentions, we construct an undirected graph where nodes are mentions and edges represent coreference links. Finally, we find connected components in the graph to obtain coreference clusters.\footnote{This method resolves to majority voting-based aggregation when the $\tau$ is set so that more than half of annotators should agree. For $\tau=N$, this method is very conservative, adding a link between two mentions only when all annotators agree unanimously. Conversely, for $\tau=1$, only a single vote is required to add a link between two mentions.} We compare aggregated annotations from \ezcoref with gold annotations across the seven datasets using B3 scores (precision, recall, and F1), as illustrated in Figure~\ref{fig:agreement_with_gold_data}. 
\paragraph{High agreement with OntoNotes, GUM, LitBank, ARRAU:} Our annotators achieve the highest precision with OntoNotes, suggesting that most of the entities identified by crowdworkers are correct for this dataset. In terms of F1 scores, the datasets which are closest to crowd annotations are GUM, LitBank, and ARRAU, all of which are annotated by experts. This result confirms that high-quality annotations can be obtained from non-experts using \ezcoref\ with minimal training.

\paragraph{Low precision with Phrase Detectives and PreCo, low recall with Quiz Bowl:}
We observe that Phrase Detectives has a very low precision compared to all other datasets, implying that crowdworkers add more links compared to gold annotations. Our qualitative analysis reveals that PD annotators miss some valid links, splitting entities which are correctly linked together by our annotators (see \autoref{table:split_entities}). Another dataset with lower precision is PreCo, which also contains many missing links. In general, we observe more actual mistakes in PreCo and PD than in the other datasets, which is not surprising as they were not annotated by experts.\footnote{That said, both PreCo and PD were additionally validated by multiple non-expert annotators.} This result is further validated by our agreement analysis of the fiction domain (\autoref{table:agreement_domain_data}), in which \ezcoref\ annotations agree far more closely with expert annotations (GUM, LitBank) than PreCo and PD. Finally, Quiz Bowl has by far the lowest recall with \ezcoref\ annotations, which is expected given the difficulty with cataphora and factual knowledge (examples (c) and (e) in Table~\ref{table:iaa_examples}).

\begin{table}[]
\tiny
\begin{tabular}{ll}
PD & \begin{tabular}[c]{@{}l@{}}Not long after \color{teal}{[}a suitor{]}\color{black} \; appeared, and as \color{teal}{[}he{]}\color{black} \; appeared to be very rich and \\the miller could see nothing in \color{teal}{[}him{]}\color{black} \; with which to find fault, he betrothed \\his daughter to \color{teal}{[}him{]}\color{black} \;. But the girl did not care for \color{teal}{[}the man{]}\color{black} \;(...). She did not \\feel that she could trust \color{brown}{[}him{]}\color{black} \;, and she could not look at \color{brown}{[}him{]}\color{black} \; nor think of \\\color{brown}{[}him{]}\color{black} \;without an inward shudder.\end{tabular} \\ \addlinespace[0.2cm]
PreCo & \begin{tabular}[c]{@{}l@{}}When I listened to the weather report, I was afraid to see
\color{violet}{[}the advertisements{]}\color{black} \;.\\ 
\color{magenta}{[}Those colorful advertisements{]}\color{black} \;always made me crazy.\end{tabular}
\end{tabular}
\vspace{-0.05in}
\caption{Cases of split entities (missing links) in annotations provided with Phrase Detectives and PreCo. Instead, our crowd annotators mark all mentions as referring to the same entity in each of these examples.}
\vspace{-0.1in}
\label{table:split_entities}
\end{table}

\begin{table}[h]
\centering \scriptsize
\begin{tabular}{llrrr}
\toprule
\multirow{2}{*}{\textbf{Domain}} & \multirow{2}{*}{\textbf{Dataset}} & \multicolumn{3}{c}{\textbf{B3}} \\ \cline{3-5} \addlinespace[0.05cm]
 &  & \textbf{Precision} & \textbf{Recall} & \textbf{F1} \\  \addlinespace[0.05cm] \hline \addlinespace[0.05cm]
\multirow{4}{*}{Fiction} & GUM & 0.982 & 0.921 & 0.950 \\ \addlinespace[0.05cm]
 & LitBank & 0.959 & 0.927 & 0.943 \\
 & PreCo & 0.805 & 0.963 & 0.877 \\
 & Phrase Detectives & 0.784 & 0.775 & 0.780 \\ \addlinespace[0.1cm]
 \bottomrule 
\end{tabular}
\vspace{-0.05in}
\caption{Agreement with existing datasets for fiction.}
\vspace{-0.1in}
\label{table:agreement_domain_data}
\end{table}



\begin{figure}[]
    \centering
   \includegraphics[width=0.48\textwidth, scale=1,clip,trim=0 0 0 0mm]{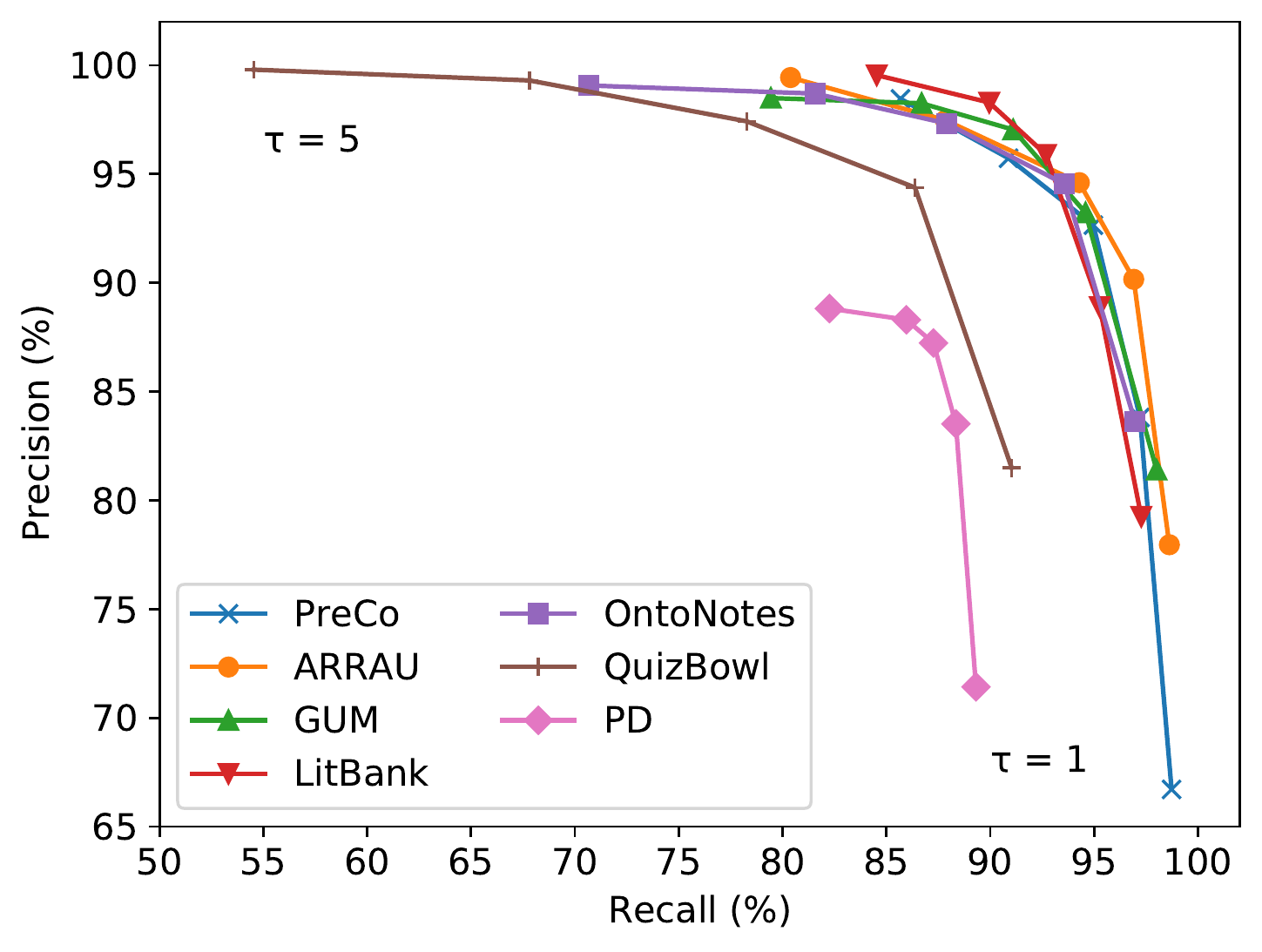}
    \vspace{-0.1in}
  \caption{Agreement with gold annotations with varying voting threshold $\tau$. $\tau=3$ is majority voting (Figure~\ref{fig:agreement_with_gold_data}). B3 scores computed with singletons included.  
  }
  \vspace{-0.1in}
  \label{fig:agreement_with_gold_votingthres}
\end{figure}

\begin{table*}[th]
\centering \tiny
\begin{tabular}{lll}
\toprule
Phenomena & \begin{tabular}[c]{@{}l@{}}Dataset \\ (Domain)\end{tabular} & Example \\ \midrule
 & \begin{tabular}[c]{@{}l@{}}LitBank\\ (Fiction)\end{tabular} & (a) {\color[HTML]{000000} \begin{tabular}[c]{@{}l@{}}A Wolf had been gorging on an animal \color{teal}{[}he{]}\color{black} \; had killed, when suddenly a small bone in the meat stuck in \color{teal}{[}his{]}\color{black} \; throat and \color{teal}{[}he{]}\color{black} \; could not swallow \color{brown}{[}it{]}\color{black}.\\ \color{teal}{[}He{]}\color{black} \; soon felt a terrible pain in \color{teal}{[}his{]}\color{black} \; throat (...) \color{teal}{[}He{]}\color{black} \; tried to induce everyone \color{teal}{[}he{]}\color{black} \; met to remove the bone. "\color{teal}{[}I{]}\color{black} \; would give anything, " said \color{teal}{[}he{]}\color{black} \;, " \\ if \color{purple}{[}you{]}\color{black} \; would take \color{brown}{[}it{]}\color{black} \; out. "\end{tabular}} \\ \addlinespace[0.1cm]
 
\multirow{-3}{*}{Pronouns} & \begin{tabular}[c]{@{}l@{}}GUM\\ (Biographies)\end{tabular} & (b) \begin{tabular}[c]{@{}l@{}}Despite Daniel's attempts at reconciliation, \color{violet}{[}his{]}\color{black} \; father carried the grudge until \color{violet}{[}his{]}\color{black} \; death. Around schooling age, \color{violet}{[}his{]}\color{black} \; father, Johann, encouraged \\ \color{violet}{[}him{]}\color{black} \; to study business (...). However, Daniel refused because \color{violet}{[}he{]}\color{black} \; wanted to study mathematics. \color{violet}{[}He{]}\color{black} \; later gave in to \color{violet}{[}his{]}\color{black} \; father's wish and studied\\ business. \color{violet}{[}His{]}\color{black} \; father then asked \color{violet}{[}him{]}\color{black} \; to study in medicine.\end{tabular} \\ \addlinespace[0.2cm]
Cataphora & \begin{tabular}[c]{@{}l@{}}QuizBowl\\

(Quizzes)\end{tabular} & (c) \begin{tabular}[c]{@{}l@{}}\color{magenta}{[}One character in this work{]}\color{black} \; is forgiven by \color{brown}{[}magenta{]}\color{black} \; wife for an affair with a governess before beginning one with a ballerina. \color{teal}{[}Another character in \\\color{teal} this work {]}\color{black} \; is a sickly, thin man who eventually starts dating a reformed prostitute, Marya Nikolaevna. In addition to \color{magenta}{[}Stiva{]}\color{black} \; and \color{teal}{[}Nikolai{]}\color{black} \;, \color{brown}{[}another \\ \color{brown} character in this work{]}\color{black} \; (...) had earlier failed in \color{brown}{[}his{]}\color{black} \;courtship of Ekaterina Shcherbatskaya.\end{tabular} \\ \addlinespace[0.2cm]

 & \begin{tabular}[c]{@{}l@{}}OntoNotes\\ (News)\end{tabular} & (d) \begin{tabular}[c]{@{}l@{}}The Soviet Union's jobless rate is soaring (...), \color{teal}{[}Pravda{]}\color{black} \; said. Unemployment has reached 27.6 \% in Azerbaijan, (...) and 16.3\% in Kirgizia, \\ \color{teal}{[}the Communist Party newspaper {]}\color{black}\;said.\end{tabular} \\ \addlinespace[0.1cm]
\multirow{-3}{*}{\begin{tabular}[c]{@{}l@{}}Factual \\ Knowledge\end{tabular}} & \begin{tabular}[c]{@{}l@{}}QuizBowl\\ (Quizes) \end{tabular} & (e) \begin{tabular}[c]{@{}l@{}}(...) \color{brown}{[} another character in this work {]}\color{black} \; (...) had earlier failed in \color{brown}{[}his{]}\color{black} \;  courtship of \color{magenta}{[}Ekaterina Shcherbatskaya{]}\color{black}. Another character in this work rejects \\ \color{magenta}{[}Ekaterina{]}\color{black} \;  before (...) moving to St. Petersburg. For 10 points name this work in which \color{brown}{[}Levin{]}\color{black} \;  marries \color{magenta}{[}Kitty{]}\color{black} \; , (...) a novel by Leo Tolstoy.\end{tabular} \\ \bottomrule
\end{tabular}
\vspace{-0.1in}
\caption{Representative examples showing unique phenomena in each dataset (coreferences are color coded).}
\vspace{-0.1in}
\label{table:iaa_examples}
\end{table*}
\paragraph{Varying the aggregation threshold $\tau$:} What is the effect of varying the aggregation threshold ($\tau$)  on precision and recall with gold annotations? Figure~\ref{fig:agreement_with_gold_votingthres} shows that the Quiz Bowl dataset has the highest drop in recall (36\% absolute drop) when increasing $\tau$ from 1 to 5.\footnote{We analyze variations in recall since it is more interpretable than precision, given that the denominator is fixed in recall with a variable number of annotators.}  This indicates that the number of unanimous clusters ($\tau=\num{5}$) is considerably lower than the total number of clusters found individually by all annotators ($\tau=\num{1}$); as such, our annotators heavily disagree about gold clusters in the QuizBowl dataset. 
We observe a similar trend in OntoNotes (26\% drop in recall), whereas Phrase Detectives has the lowest drop in recall (\num{0.07}) with the increase in the number of annotators, 
which is expected since Phrase Detectives is crowdsourced.

\subsection{What domains are most suitable for crowdsourcing coreference?}
Which domains yield the highest inter-annotator agreement (IAA) between our crowdworkers? 

\begin{table}[h]
\centering \scriptsize
\begin{tabular}{lllllll}
\toprule
\textbf{Fiction} & \textbf{Bio} & \textbf{Opinion} & \textbf{Web} & \textbf{News} & \textbf{Wiki} & \textbf{Quiz} \\ \midrule
 72.6             & 72.4         & 69.5             & 65.9         & 62.3          & 61.8          & 59.7          \\ \bottomrule
\end{tabular}
 \caption{Inter Annotator Agreement (B3 \%) across different domains. B3 scores are computed in accordance with the CoNLL script~\cite{pradhan-etal-2014-scoring}, excluding singletons. Bio (Biographies); Wiki (Wikipedia).}
\vspace{-0.1in}
 \label{table:iaa_agreement}
\end{table}
We use the B3 metric\footnote{We also computed Krippendorff's $\alpha$ for inter-annotator agreement and obtained similar results.} \cite{bagga-1998-algorithms} to compute IAA for each domain, excluding singletons\footnote{The agreement including singletons is substantially higher. The exact numbers are provided in Appendix \ref{sec:iaa_appendix_singletons}.} (see Table~\ref{table:iaa_agreement}).
We obtain the highest agreement on fiction (72.6$\%$) and biographies (72.4$\%$). This is because both domains contain a high frequency of pronouns (see examples \emph{a} and \emph{b} in Table~\ref{table:iaa_examples}), which our annotators found easier to annotate.
We also observe that the fiction domain contains many well-known children stories (e.g., Little Red Riding Hood) that are likely familiar to our annotators, which may have made them easier to annotate.
Annotators have the least agreement on Quiz Bowl coreference (59.73$\%$), as this dataset is rich in challenging cataphoras (example \emph{c} in Table~\ref{table:iaa_examples}) and often require world knowledge about books, characters, and authors to identify coreferences (example \emph{e} in Table~\ref{table:iaa_examples}).

\subsection{Qualitative analysis}
\label{sec:qualitative}


To better understand the differences in annotation quality, we conduct a manual analysis\footnote{Conducted by a linguist who studied annotation guidelines of all datasets.} of all 240 passages in our experiment, comparing our \ezcoref\ annotations to gold annotations from each dataset. Specifically, we look at each link that was annotated by our workers but not in the gold data, or vice versa. For each link, we determine whether crowd or the gold annotations contained a mistake, or whether the discrepancy is reasonable under specific guidelines. We find that \ezcoref\ annotations contain fewer mistakes than non-expert annotated datasets such as PreCo and PD, but there are almost twice as many mistakes as those of expert datasets such as OntoNotes and GUM, and seven times as many mistakes as those in the esoteric Quiz Bowl dataset (\autoref{table:mistakes} in Appendix).

\paragraph{Qualitative examples of disagreements and deviations from expert guidelines:}
As in~\citet{Poesio2005AnnotatingA}, we identify cases of genuine ambiguity, where a mention can refer to two different antecedents. The first row of Table~\ref{table:fog_you_examples} shows an example from Dickens' \emph{Bleak House}, where the pronoun ``it'' could reasonably refer to either the ``fog'' or the ``river.'' Our annotators have high disagreement on this link, which is understandable given the literary analysis of~\citet{szakolczai2016novels} who interprets the ambiguity of this pronoun as Dickens' way to show indeterminacy attributed to elements in the scene.\footnote{In LitBank, the source of this passage, the pronoun ``it'' is annotated as referring to the ``river'' as only ``river'' is a potential markable per entity restriction (selects ACE entities only).}

We observe that generic mentions, especially generic pronouns, are almost always annotated as coreferring by our annotators. The second row of Table~\ref{table:fog_you_examples} shows one example of such a case, where annotators unanimously connected all instances of generic ``you.'' While generic pronouns are usually regarded as non-referring \cite{huddleston2002the_you_nonrefering}, they retain something of their specific quality as personal pronouns \cite{quirk1985a_you_retained}.
Furthermore, \citet{Orvell2020} demonstrate that the use of generic ``you'' promotes the resonance between people and ideas. By using the same linguistic form (``you''), often used to refer to the addressee, one invites the addressee to consider how the situation refers to them, and in this sense, the generic ``you'' is being perceived as referring to the same entity, the reader, or, in our case, the annotator. 
Finally, while datasets tend to treat copulae and appositive constructions identically and annotate them in a similar way, our annotators intuitively annotate them differently. 
Although they almost always mark noun phrases in appositive constructions as coreferents, the noun phrases in copulae are linked by majority vote only in $\sim$ 35\% of cases.


\begin{table}[]
\tiny
\begin{tabular}{ll}

Ambiguity & \begin{tabular}[c]{@{}l@{}}\color{teal}{[}Fog{]}\color{black} \; everywhere. \color{teal}{[}Fog{]}\color{black}  \; up \color{brown}{[}the river{]}\color{black} \;, where \color{blue}{[}it{]}\color{black} \; flows among green \\ aits and meadows; \color{teal}{[}fog{]}\color{black}  \;  down \color{brown}{[}the river{]}\color{black} \;, where \color{blue}{[}it{]}\color{black} \; rolls defiled  among \\the tiers of shipping and the waterside pollutions of a great (and dirty) city. \\ -  Charles Dickens, \textit{Bleak House} \end{tabular}\\ \addlinespace[0.2cm]

Generic & \begin{tabular}[c]{@{}l@{}}Please , Ma’am , is this New Zealand or Australia? ( and she tried to \\ curtsey as she spoke -- fancy CURTSEYING as \color{violet}{[}you{]}\color{black} \;’re falling \\ through the air! Do \color{violet}{[}you{]}\color{black} \; think \color{violet}{[}you{]}\color{black} \; could manage it?) \\
- Lewis Carroll, \textit{Alice in Wonderland} \end{tabular}

\end{tabular}
\vspace{-0.1in}
\caption{Examples of genuine ambiguity and generic ``you'' observed in our data.}
\vspace{-0.1in}
\label{table:fog_you_examples}
\end{table}
\section{Conclusion}
We investigate whether it is feasible to crowdsource coreference annotations by providing \emph{minimal} guidelines to non-expert annotators, thereby saving annotator and researcher efforts. Concretely, we develop a crowdsourced coreference platform called \ezcoref and use it to re-annotate 240 passages from seven existing English coreference datasets. Our crowd workers agree with expert annotations even without extensive training, signifying that high-quality data can be obtained via crowdsourcing with minimalistic guidelines. We also observe crowd deviations from expert guidelines on linguistic phenomena such as general pronouns, appositives, and copulae. 
We hope our observations will inform guideline creation for future coreference annotation efforts.
\section{Limitations}
\label{sec:limitations}
We list some of the limitations of our study which researchers and practitioners would hopefully benefit from when interpreting our analysis. Firstly, our analysis is only applicable to the English language and how native English speakers understand coreferences. In this work, we have taken a step towards building a framework to facilitate the comparison of the crowd and expert annotations, and the variations observed in non-native speakers should be explored in future studies. Secondly, as a result of resource constraints, we limited ourselves to one set of guidelines and compared crowd annotations under these guidelines with expert annotations. Understanding the effects of various guidelines on annotator behavior is left for future research. Thirdly, even the best automatic mention detection algorithm could have errors, especially when tested out-of-domain. Some of the proposed solutions are to directly crowdsource mentions or verify the automatically identified mentions via crowdsourcing~\cite{madge2019progression}, which can be utilized for future collection of high-quality corpora.

\section{Ethics Statement}
\label{sec:ethics}

The data collection protocol was approved by the coauthors' institutional review board. All annotators were presented with a consent form (mentioned below) prior to the annotation. They were also informed that only satisfactory performance on the screening example will allow them to take part in the annotation task. All data collected during the tutorial and annotations (including annotators` feedback and demographics) will be released anonymized. 
We also ensure that the annotators receive at least \$13.50 per hour. Since base compensation is per unit of work, not by time (the standard practice on Amazon Mechanical Turk), we add bonuses for workers whose speed caused them to fall below that hourly rate.

\paragraph{Consent}
Before participating in our study, we requested every annotator to provide their consent. The annotators were informed about the purpose of this research study, any risks associated with it, and the qualifications necessary to participate. The consent form also elaborated on task details describing what they will be asked to do and how long it will take. The participants were informed that they could choose as many documents as they would like to annotate (by accepting new Human Intelligence Tasks at AMT) subject to availability, and they may drop out at any time. Annotators were informed that they would be compensated in the standard manner through the Amazon Mechanical Turk crowdsourcing platform, with the amount specified in the Amazon Mechanical Turk interface. As part of this study, we also collected demographic information, including their age, gender, native language, education level, and proficiency in the English language. We ensured our annotators that the collected personal information would remain confidential in the consent form.
\section*{Acknowledgements}


We are very grateful to the crowd annotators on AMT
for participating in our annotation tasks and providing positive reviews. We are grateful to Kavya Jeganathan, Abe Handler, Neha Kennard, Timothy O'Gorman, Nishant Yadav, Anna Rogers, and the UMass NLP group for several useful discussions during the course of the project. We also thank Massimo Poesio for sharing the GNOME portion of ARRAU dataset. 
This material is based upon work supported by National Science Foundation awards 1925548, 1814955, and 1845576, and a Google PhD Fellowship awarded to KK. 

\bibliographystyle{acl_natbib}
\bibliography{bib/anthology, bib/custom}

\clearpage
\setcounter{table}{0}
\renewcommand{\thetable}{A\arabic{table}}
\appendix
\section{Appendix}
\label{sec:appendix}
\subsection{Details of our crowdsourced data}
\autoref{table:datasets-description-detailed} mentions all datasets that we re-annotate in this work with their breakdown based on domains, number of documents, passages, tokens and mentions annotated.
\label{sec:crowd_data}





\begin{table}[h]
\centering \tiny
\begin{tabular}{llrrrr}
\hline
\textbf{Dataset} & \textbf{Domain} & \multicolumn{1}{l}{\textbf{\#Docs}} & \multicolumn{1}{l}{\textbf{\#Passages}} & \multicolumn{1}{l}{\textbf{\#Tokens}} & \multicolumn{1}{l}{\textbf{\#Mentions}} \\ \hline
\multirow{3}{*}{OntoNotes} & News & 6 & 30 & 4923 & 1365 \\
 & Weblogs & 5 & 20 & 3452 & 1001 \\
 & Opinion & 12 & 20 & 3861 & 1157 \\
LitBank & Fiction & 4 & 30 & 5455 & 1494 \\
QuizBowl & Quizzes & 20 & 20 & 3304 & 1083 \\
ARRAU & News & 3 & 20 & 3336 & 885 \\
\multirow{2}{*}{GUM} & Biographies & 4 & 20 & 3422 & 1119 \\
 & Fiction & 4 & 20 & 3299 & 1008 \\
\multirow{2}{*}{\begin{tabular}[c]{@{}l@{}}Phrase \\ Detectives\end{tabular}} & Wikipedia & 7 & 20 & 3509 & 1003 \\
 & Fiction & 4 & 20 & 4007 & 1063 \\
\multirow{4}{*}{PreCo} & Opinion & 7 & 9 & 1692 & 495 \\
 & News & 4 & 8 & 1318 & 369 \\
 & Fiction & 2 & 2 & 378 & 105 \\
 & Biographies & 1 & 1 & 152 & 53 \\ \hline
Total & All & 83 & 240 & 42108 & 12200 \\ \hline
\end{tabular}
\caption{All datasets analyzed in this work with their breakdown based on domains, number of documents, passages, tokens and mentions annotated.}
\label{table:datasets-description-detailed}
\end{table}

\subsection{Manual Qualitative Analysis}

\begin{table}[h]
\centering \scriptsize
\begin{tabular}{lrr}
\toprule
\multirow{2}{*}{\textbf{Dataset}} & \multirow{2}{*}{\textbf{Mistakes (our)}} & \multirow{2}{*}{\textbf{Mistakes (gold)}} \\
                     &    &    \\ \midrule
\textbf{PD (silver)} & 22 & 76 \\
\textbf{PreCo}       & 12 & 33 \\
\textbf{GUM}         & 48 & 25 \\
\textbf{OntoNotes}   & 81 & 49 \\
\textbf{ARRAU}       & 33 & 16 \\
\textbf{LitBank}     & 21 & 13 \\
\textbf{QuizBowl}    & 67 & 10 \\ \bottomrule
\end{tabular}
\caption{Number of mistakes in our crowd annotations vs. gold datasets, obtained through a manual analysis.}
\label{table:mistakes}
\end{table}

\subsection{Detailed Mention Detection Algorithm}
\label{sec:md_algo}
\begin{itemize}
    \item We identify all noun phrases using the Stanza dependency parser~\cite{qi2020stanza}. For each word with a noun-related part-of-speech tag,\footnote{Pronouns, nouns, proper nouns, and numbers.} we recursively traverse all of its children in the dependency graph until a dependency relation is found in a \texttt{whitelist}.\footnote{The \texttt{whitelist} includes all multi-word expression relations (i.e., compound, flat, and fixed) and modifier relations (i.e., determiners, adjectival modifiers, numeric modifiers, nominal modifiers, and possessive nominal modifiers).} The maximal span considered as a candidate mention thus covers all words related by relations in the \texttt{whitelist}.
    %
    
    
    \item Possessive nominal modifiers are also considered as candidate mentions. For instance, in the sentence ``Mary's book is on the table,'' we consider both ``Mary'' and ``Mary's book'' as mentions.
    
    \item Modifiers that are proper nouns in a multi-word expression are considered as mentions. For instance, in ``U.S. foreign policy,'' the modifier  ``U.S.'' is also considered as a mention. 
    
    \item All conjuncts, including the headword and other words depending on it via the conjunct relation, are considered mentions in a coordinated noun phrase. For instance, in the sentence, ``John, Bob, and Mary went to the party.'', the detected mentions are ``John,'' ``Bob,'' ``Mary,'' and the coordinated noun phrase ``John, Bob, and Mary.''
    
    \item Finally, we remove mentions if a larger mention with the same headword exists. We allow nested spans (e.g., \textcolor{red}{\textbf{[}}\textcolor{teal}{\textbf{[}}my\textcolor{teal}{\textbf{]}} hands\textcolor{red}{\textbf{]}}) but merge any intersecting spans into one large span (e.g, \textcolor{red}{\textbf{[}}western \textcolor{teal}{\textbf{[}}Canadian\textcolor{red}{\textbf{]}} province\textcolor{teal}{\textbf{]}} is merged into \textbf{[}western Canadian province\textbf{]}).
\end{itemize}

\subsection{Inter-Annotator Agreement Among Our Annotators Across Domains}
\begin{figure}[h]
    \centering
  \includegraphics[width=0.5\textwidth, scale=1,clip,trim=0 0 0 0mm]{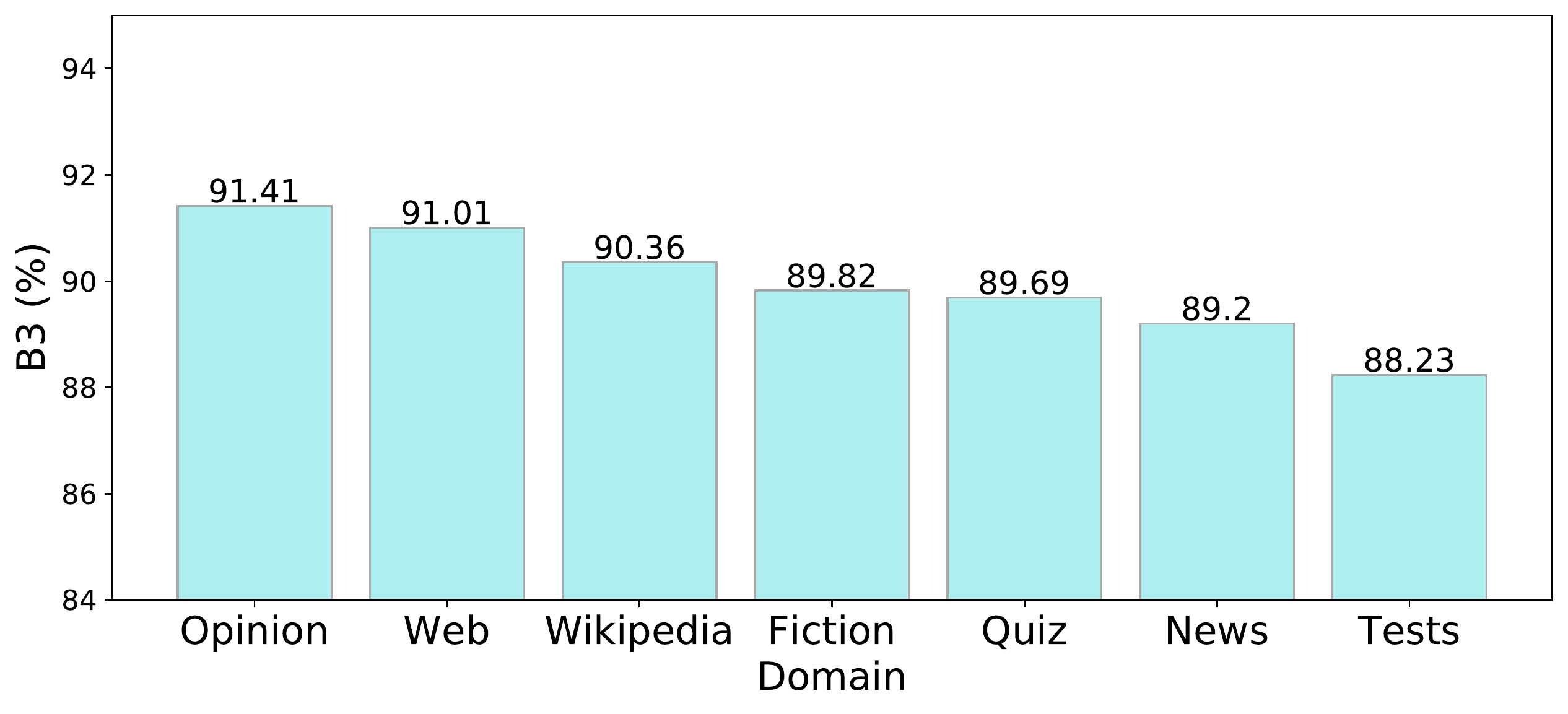}
  \caption{Inter Annotator Agreement across different domains. B3 scores with Singletons included.}
  \label{fig:IAA_datawise_include_singletons}
\end{figure}

Figure~\ref{fig:IAA_datawise_include_singletons} illustrates agreement among our annotators computed with B3 scores including singletons.

\label{sec:iaa_appendix_singletons}

\subsection{Another illustrative example}
An example of a single sentence annotated by two datasets, OntoNotes and ARRAU. These annotations differ widely from each other in kinds of mentions and links between mentions. 

\begin{quote}
\small
\textbf{OntoNotes}: {\color{red} \texttt{[}} Lloyd's, once a pillar of {\color{blue} \texttt{[}} the world insurance market {\color{blue} \texttt{]e1}}, {\color{red} \texttt{]e2}} is being shaken to {\color{red} \texttt{[}} its {\color{red} \texttt{]e2}} very foundation.\vspace{0.2cm}\\
\textbf{ARRAU}: {\color{red} \texttt{[}} Lloyd's, once {\color{violet}\texttt{[}} a pillar of {\color{blue}\texttt{[}} the world {\color{teal}\texttt{[}} insurance {\color{teal}\texttt{]e3}} market \texttt{{\color{blue}]e2} {\color{violet}]eS1} {\color{red} ]e1}}, is being shaken to \texttt{{\color{brown}[} {\color{red}[}} its {\color{red}\texttt{]e1}} very foundation {\color{brown} \texttt{]eS2}}.
\end{quote}

\clearpage
\begin{table*}[t!]
\tiny
\begin{center}
\begin{tabular}{ lcccccccccc} 
 \toprule
System & \begin{tabular}[c]{@{}l@{}}Annotate\\all clusters\end{tabular} & \begin{tabular}[c]{@{}l@{}}Pre-identified \\Mentions\end{tabular} & \begin{tabular}[c]{@{}l@{}}Open\\Source\end{tabular} & Webapp & \begin{tabular}[c]{@{}l@{}}Coref\\only\end{tabular} & \begin{tabular}[c]{@{}l@{}}Keyboard\\and Mouse\end{tabular} & \begin{tabular}[c]{@{}l@{}}MTurk\\Tested\end{tabular} & \begin{tabular}[c]{@{}l@{}}Non-expert\\Terminology\end{tabular} & \begin{tabular}[c]{@{}l@{}}Nested Span\\Support\end{tabular} & \begin{tabular}[c]{@{}l@{}}Interactive\\Tutorial\end{tabular} \\
\midrule
\citet{stenetorp-etal-2012-brat} & \checkmarkcol & \xmarkcol & \checkmarkcol & \checkmarkcol & \xmarkcol & \xmarkcol & \xmarkcol & \checkmarkcol & \xmarkcol$^*$ & \xmarkcol \\ \addlinespace[0.1cm]
\citet{widlcher-2012-glozz} & \checkmarkcol & \xmarkcol & \xmarkcol & \xmarkcol & \xmarkcol & \xmarkcol & \xmarkcol & \xmarkcol & \checkmarkcol & \xmarkcol \\ \addlinespace[0.1cm]
\citet{landragin-etal-2012-analec} & \checkmarkcol & \xmarkcol &  \checkmarkcol & \xmarkcol & \xmarkcol & \xmarkcol & \xmarkcol & \xmarkcol & \checkmarkcol & \xmarkcol\\ \addlinespace[0.1cm]
\citet{yimam-etal-2013-webanno} & \checkmarkcol & \xmarkcol & \checkmarkcol & \checkmarkcol & \xmarkcol & \xmarkcol & \xmarkcol$^*$ & \xmarkcol & \checkmarkcol & \xmarkcol \\ \addlinespace[0.1cm]
\citet{poesio-etal-2013-phrase} & \xmarkcol & \checkmarkcol & \xmarkcol & \checkmarkcol & \checkmarkcol & \xmarkcol & \xmarkcol & \checkmarkcol &\checkmarkcol & \checkmarkcol \\ \addlinespace[0.1cm]
\citet{girardi-etal-2014-cromer} & \xmarkcol & \xmarkcol & \checkmarkcol & \checkmarkcol & \checkmarkcol & \xmarkcol & \xmarkcol & \xmarkcol & \xmarkcol & \xmarkcol \\ \addlinespace[0.1cm]
\citet{kopec-2014-mmax2} & \checkmarkcol & \xmarkcol & \checkmarkcol & \xmarkcol & \checkmarkcol & \xmarkcol & \xmarkcol & \xmarkcol & \checkmarkcol & \xmarkcol \\ \addlinespace[0.1cm]
\citet{guha-etal-2015-removing} & \checkmarkcol & \xmarkcol & \checkmarkcol & \checkmarkcol & \checkmarkcol & \checkmarkcol & \xmarkcol & \checkmarkcol & \checkmarkcol & \xmarkcol \\ \addlinespace[0.1cm]
\citet{oberle-2018-sacr} & \checkmarkcol & \xmarkcol & \checkmarkcol & \checkmarkcol & \checkmarkcol & \xmarkcol & \xmarkcol & \xmarkcol & \checkmarkcol & \xmarkcol \\ \addlinespace[0.1cm]
\citet{reiter-2018-corefannotator} & \checkmarkcol & \xmarkcol & \checkmarkcol & \xmarkcol & \checkmarkcol & \xmarkcol & \xmarkcol & \xmarkcol & \checkmarkcol & \xmarkcol \\ \addlinespace[0.1cm]
\citet{bornstein-2020-corefi} & \checkmarkcol & \checkmarkcol & \checkmarkcol  & \checkmarkcol & \checkmarkcol & \xmarkcol & \checkmarkcol & \xmarkcol & \xmarkcol & \checkmarkcol \\ \addlinespace[0.1cm]
\midrule
\ezcoref (this work) & \checkmarkcol & \checkmarkcol & \checkmarkcol$^*$  & \checkmarkcol & \checkmarkcol & \checkmarkcol  & \checkmarkcol & \checkmarkcol & \checkmarkcol & \checkmarkcol \\
\bottomrule
\end{tabular}
\end{center}
\caption{A comparison of different coreference annotation tools. (* --- \ezcoref~code will be open-sourced upon paper publication; \citet{stenetorp-etal-2012-brat} did not implement nested spans originally, but later added them with limited functionality. \citet{yimam-etal-2013-webanno} have APIs for CrowdFlower integration, but suggest expert annotators.)}
\label{tab:tools-prior-work-compare}
\end{table*}


\begin{table*}[h]
\centering \scriptsize
\begin{tabular}{l}
\hline \addlinespace[0.2cm]
\multicolumn{1}{c}{\textbf{Tutorial feedback from our crowd annotators}} \\ \addlinespace[0.2cm] \hline \addlinespace[0.2cm]
1. \begin{tabular}[c]{@{}l@{}}This was a really interesting task. The tutorial was very clear and easy to understand. I think it was very helpful when\\  I completed the final passage.\end{tabular} \\    \addlinespace[0.2cm]
2. Very great tutorial, I loved how it walked me through each and every step making sure I understood. \\   \addlinespace[0.2cm]
3. \begin{tabular}[c]{@{}l@{}}excellent interface and very precise instructions!  out of curiousity, what is the time-frame and scale for this project? \\ several weeks? months? hundreds or thousands of hits?  I have a ton of projects during the autumn normally but will\\  definitely make time for this if it's going to be around for more than a day or two.  Looking forward to working with\\  you folks if possible!\end{tabular} \\   \addlinespace[0.2cm]
4. I actually enjoyed this. Thank you for the opportunity. \\   \addlinespace[0.2cm]
5. it was interesting a bit difficult but overall gave a lot of feedback necessary to do a good job. \\   \addlinespace[0.2cm]
6. \begin{tabular}[c]{@{}l@{}}I loved the tutorial and the layout.  I am still a little bit unsure about a couple of the entities and hope I got it right.  \\ For example: would 'legs' be in 'his' because it refers to that person? I wasn't sure and made them separate.\end{tabular} \\   \addlinespace[0.2cm]
7. \begin{tabular}[c]{@{}l@{}}I loved how this tutorial was set up. It was easy to use and made me very interested in doing the actual HITs.\\ It would have been nice to be able to print out a quick reference guide or something, so we could refer to the \\ instructions from before while we completed the final task. I don't think it would be needed for very long after \\ starting the real HITs, but it would still be nice to have.\end{tabular} \\   \addlinespace[0.2cm]
8. \begin{tabular}[c]{@{}l@{}}On the last test section, there was no place for feedback. There was a section that said ""it was getting dark"" \\ ""It was getting late"" Both of those refer to a time of day, but one is light, one is the hour, so I marked them \\ as different. Not sure of how broad or narrow we need to be when justifying ""same"" entities, as there is an \\ argument either way.\end{tabular} \\   \addlinespace[0.2cm]
9. I just wanted to say that I really appreciated how efficiently put together and clear this tutorial was. \\   \addlinespace[0.2cm]
10. This was a unique task. Thank you. \\   \addlinespace[0.2cm]
11. \begin{tabular}[c]{@{}l@{}}I feel much better with the help and feedback. It was interesting and definitely way different in a good way than \\ the usual survey. I did my best and I hope I did well enough. Keep safe and Happy Holidays no matter what happens.\end{tabular} \\ \addlinespace[0.2cm] \hline
\end{tabular}
\caption{Some of the comments received from our annotators after completing the tutorial. We received overwhelmingly positive feedback; annotators sometimes also mentioned cases they found confusing.}
\label{table:tutorial_feedback}
\end{table*}


\begin{figure*}[t!]
    \centering
  \includegraphics[width=\textwidth, scale=1,clip,trim=0 0 0 0mm]{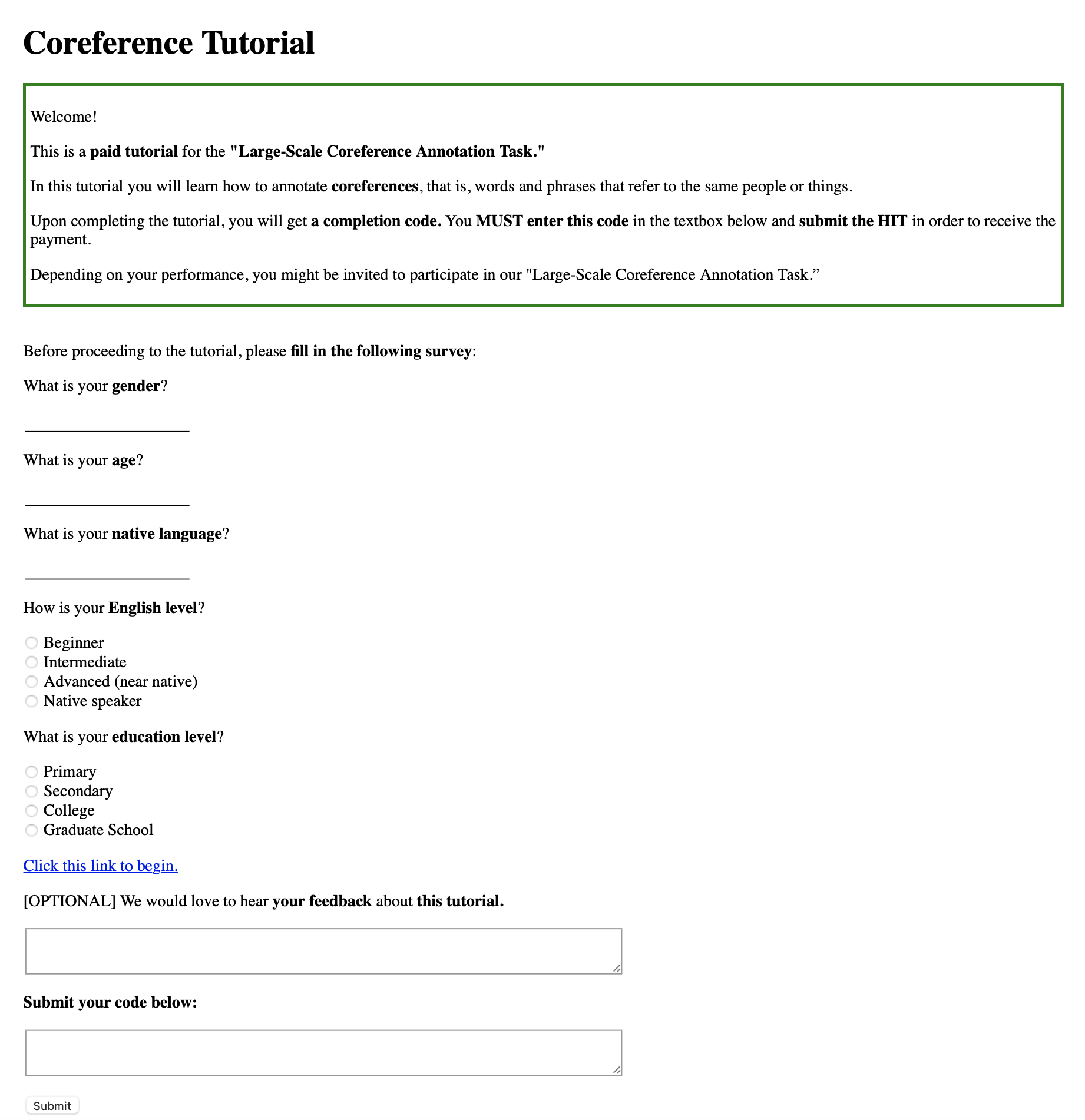}
  \caption{Screenshot of tutorial task invitation on AMT with detailed instructions.}
  \label{fig:tutorial_4}
\end{figure*}

\begin{figure*}[t!]
    \centering
  \includegraphics[width=0.5\textwidth, scale=0.1,clip,trim=0 0 0 0mm]{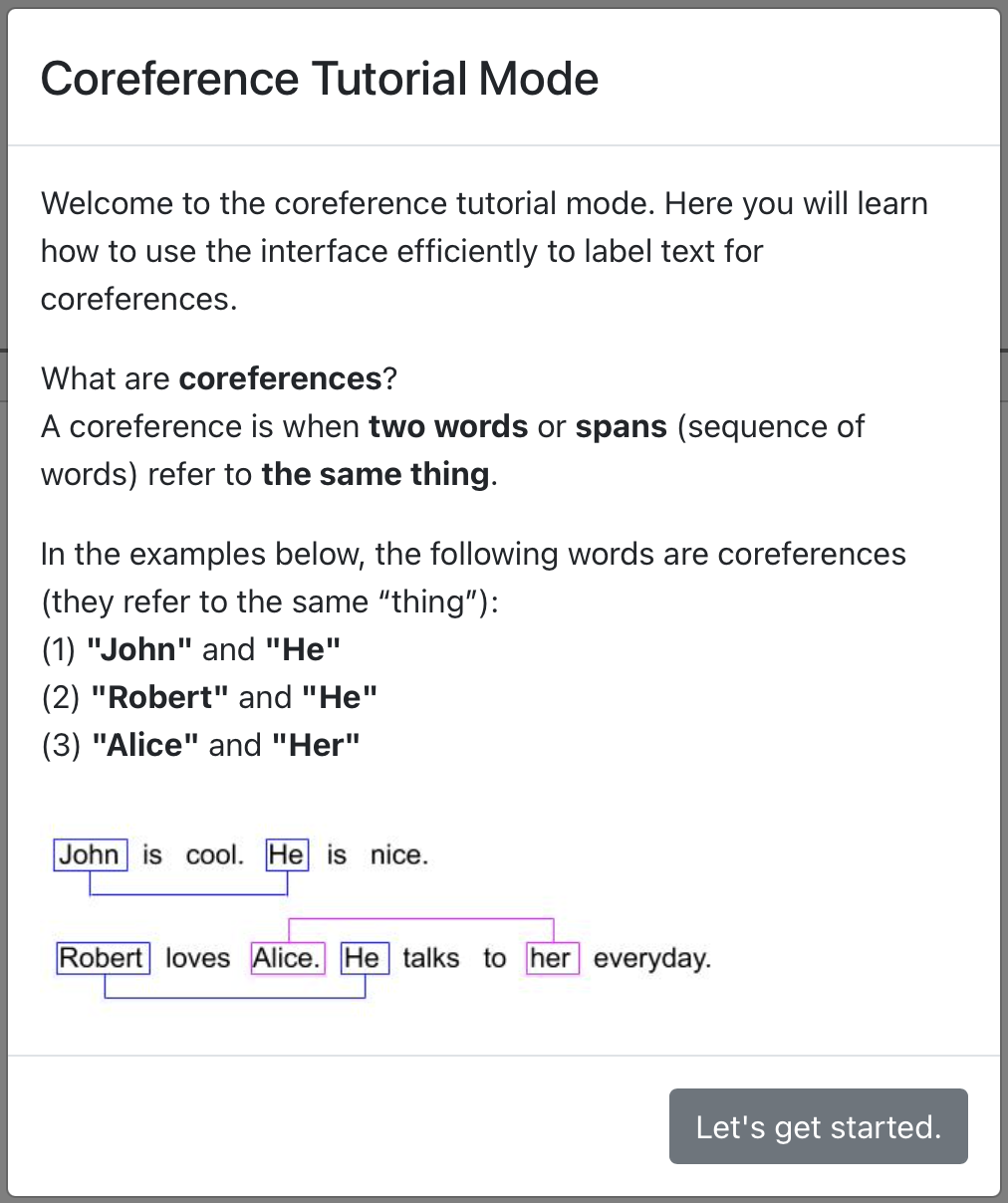}
  \caption{ Tutorial Interface (Introductory prompt) }
  \label{fig:tutorial_1}
\end{figure*}

\begin{figure*}[t]
    \centering
  \includegraphics[width=\textwidth, scale=1,clip,trim=0 0 0 0mm]{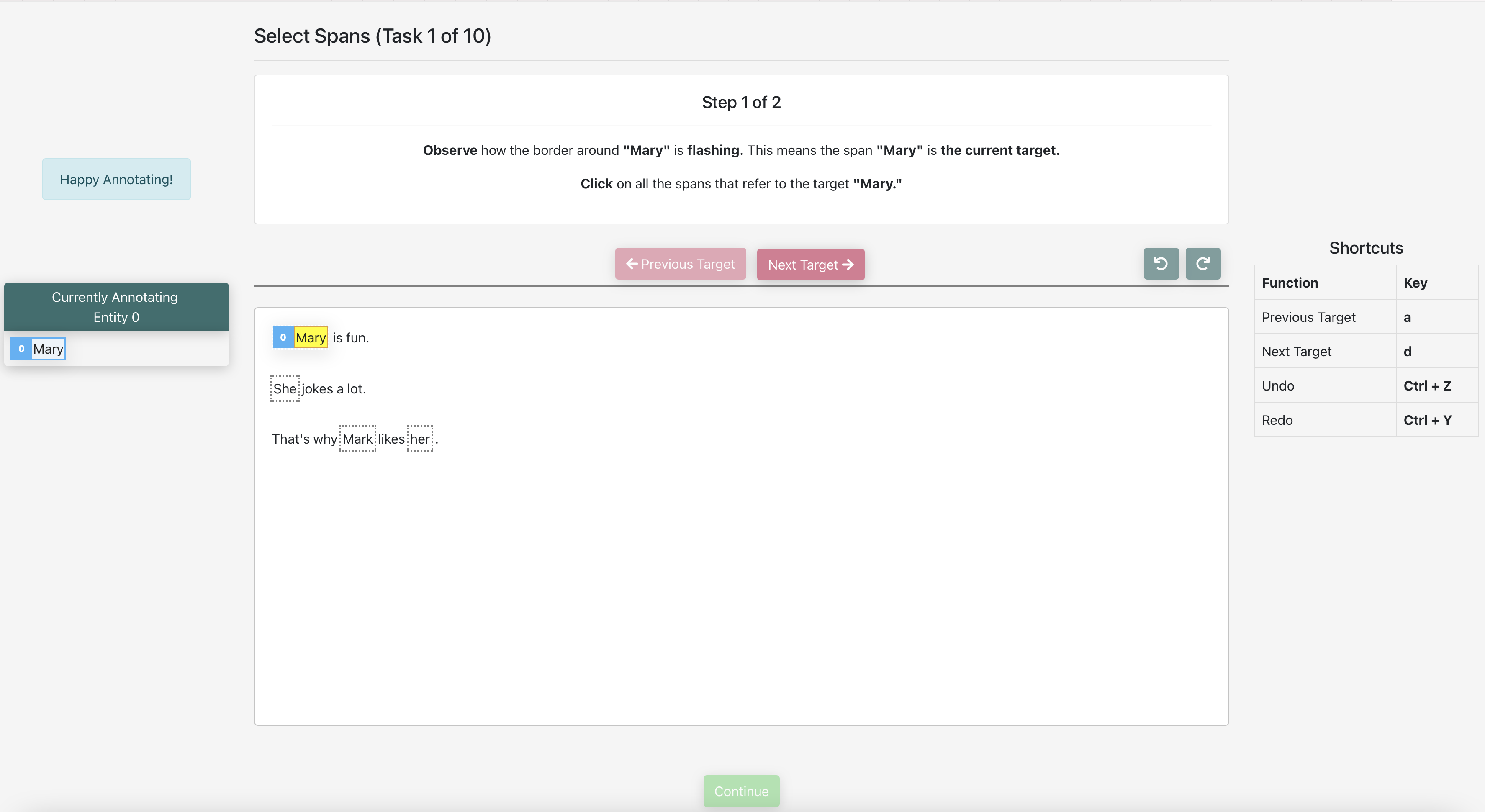}
  \caption{Tutorial interface: A sample prompt teaching tool functionality.}
  \label{fig:tutorial_2}
\end{figure*}

\begin{figure*}[t]
    \centering
  \includegraphics[width=\textwidth, scale=1,clip,trim=0 0 0 0mm]{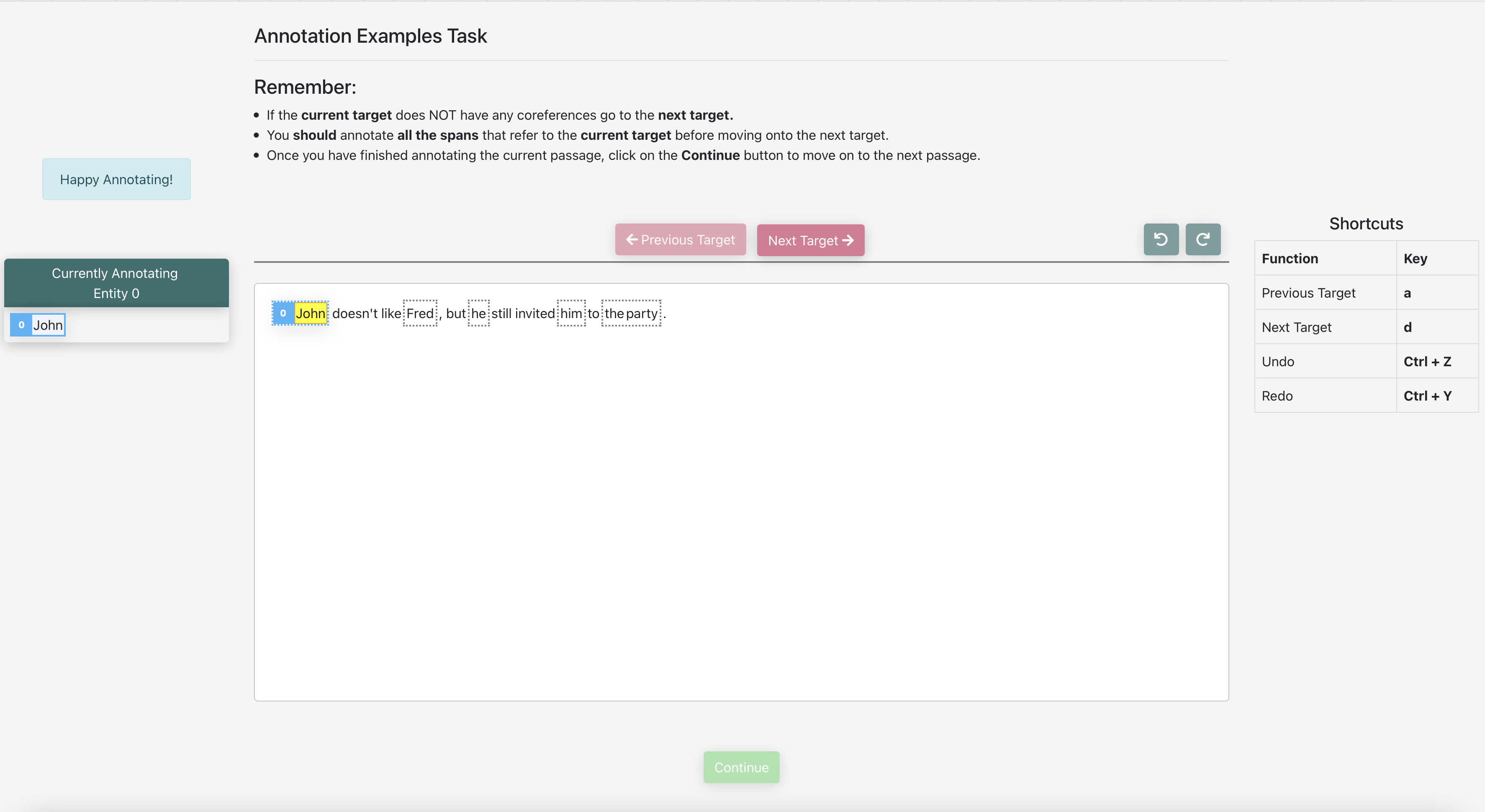}
  \caption{Tutorial interface: A sample prompt teaching basic coreferences.}
  \label{fig:tutorial_3}
\end{figure*}

\begin{figure*}[t]
    \centering
  \includegraphics[width=\textwidth, scale=1,clip,trim=0 0 0 0mm]{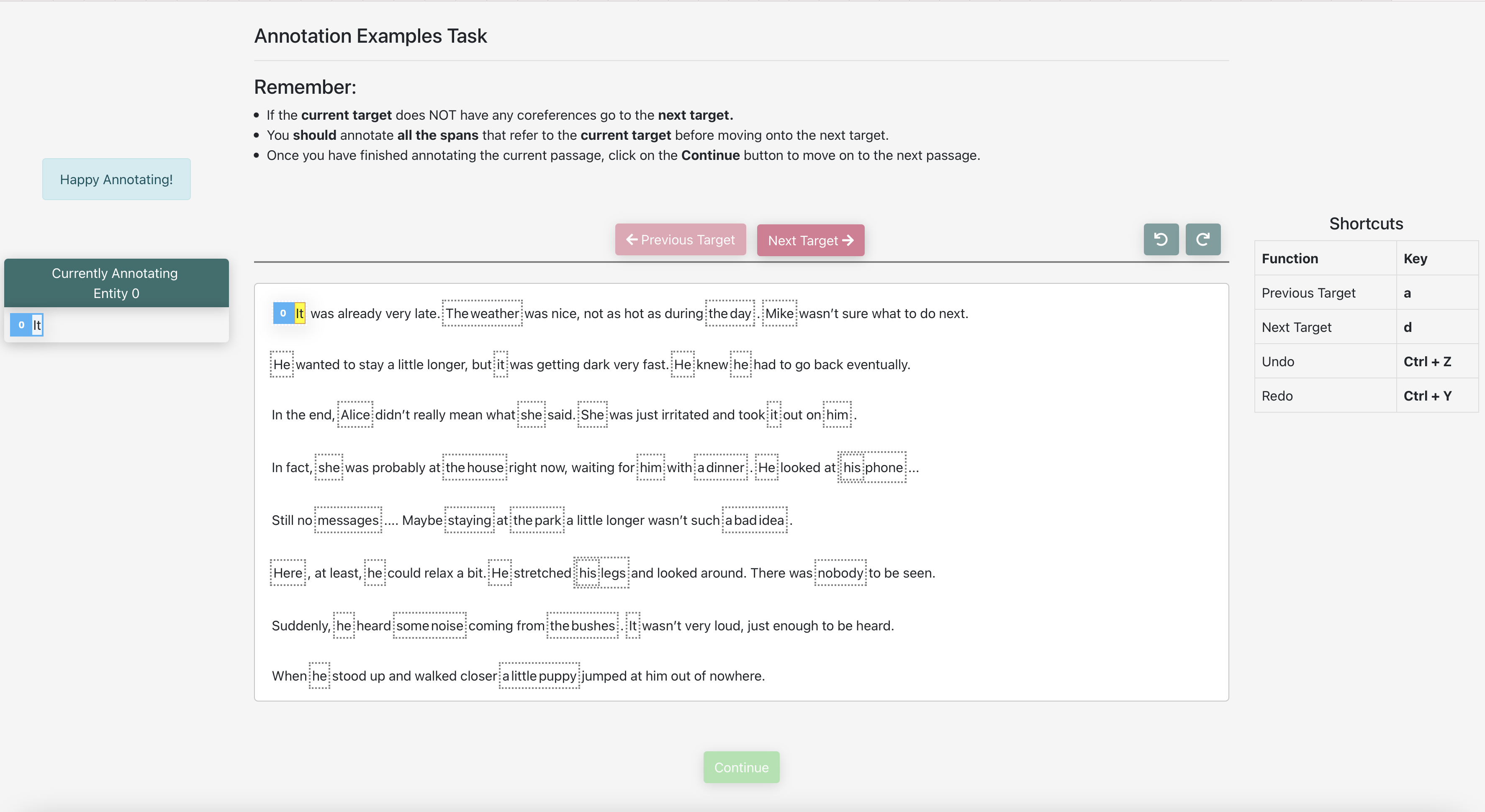}
  \caption{Tutorial interface: quality control example.}
  \label{fig:tutorial_4}
\end{figure*}

\begin{figure*}[t]
    \centering
  \includegraphics[width=\textwidth, scale=1,clip,trim=0 0 0 0mm]{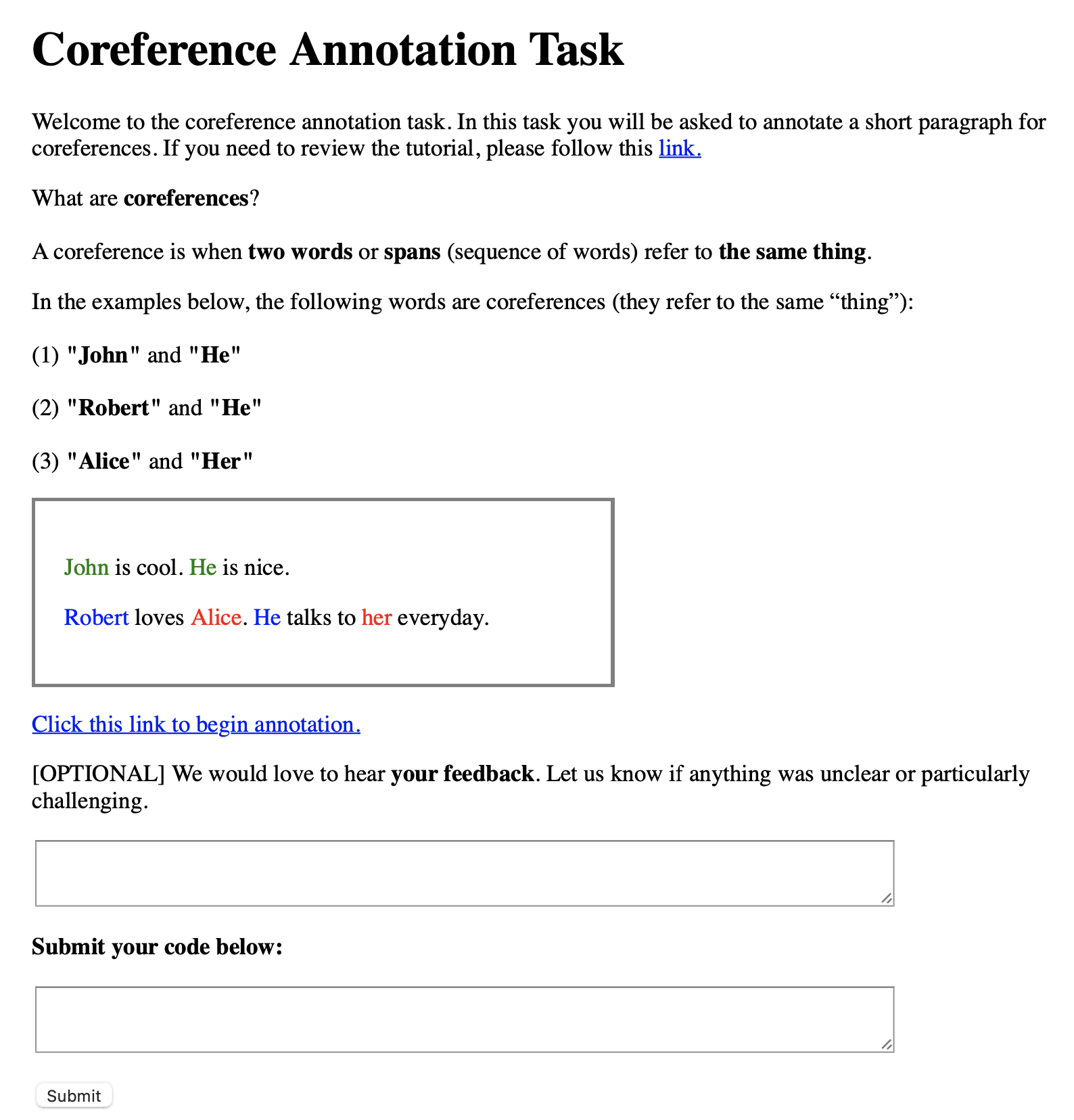}
  \caption{Annotation task invite on AMT with detailed instructions}
  \label{fig:tutorial_4}
\end{figure*}
\clearpage



\end{document}